\pdfoutput=1

\documentclass[11pt]{article}
\usepackage[]{acl}
\usepackage{listings}
\usepackage{caption} 
\usepackage{times}
\usepackage{latexsym}
\usepackage{graphicx}
\usepackage{multirow}
\usepackage{amsmath}

\usepackage{booktabs}
\usepackage{soul}
\usepackage{arydshln}
\usepackage{url}
\usepackage{minitoc}
\usepackage{tocbibind}
\usepackage{titletoc}
\usepackage{etoc}
\usepackage{wrapfig}
\usepackage{cleveref}
\usepackage{enumitem}
\usepackage{float}
\usepackage{placeins}
\usepackage{hyperref}
\usepackage[T1]{fontenc}

\usepackage[utf8]{inputenc}

\usepackage{microtype}
\usepackage{xspace}      
\usepackage{ulem}

\newcommand{\system}[1]{\textsc{#1}\xspace}  
\newcommand{\llamaTwoThirteenB}{\system{Llama2-13b-chat}}  
\newcommand{\llamaTwoSeventyB}{\system{Llama2-70b-chat}}
\newcommand{\llamaThreeSeventyB}{\system{Llama-3-70b-chat}}
\newcommand{\gptThreeFive}{\system{GPT-3.5-turbo}}
\newcommand{\codeLlamaSevenB}{\system{CodeLlama-7b}}
\newcommand{\llamaThreeEightB}{\system{Meta-Llama-3-8B}}

\newcommand{\olmo}{\system{Olmo-7b}}
\newcommand{\olmosft}{\system{Olmo-7b-SFT}}
\newcommand{\starcoder}{\system{Starcoder-15.5b}}
\newcommand{\octocoder}{\system{Octocoder-15.5b}}

\newcommand{\SCAR}{\system{SCAR}}
\newcommand{\SCARID}{\system{SCAR(ID)}}
\newcommand{\SCAROOD}{\system{SCAR(OOD)}}
\newcommand{\AlpaGasus}{\system{AlpaGasus}}
\newcommand{\HFR}{\system{HFR}}
\newcommand{\Superfiltering}{\system{Superfiltering}}

\newcommand{\Random}{\system{Random}}
\newcommand{\Perplexity}{\system{Perplexity}}
\newcommand{\Diversity}{\system{Diversity}}
\newcommand{\Longest}{\system{Longest}}

\newcommand{\dataset}[1]{\textsc{#1}\xspace}
\newcommand{\HumanEval}{\dataset{HumanEval}}
\newcommand{\AlpacaEval}{\dataset{AlpacaEval}}
\newcommand{\StackExchange}{\dataset{StackExchange}}
\newcommand{\LIMA}{\dataset{LIMA}}
\newcommand{\MultiPLE}{\dataset{MultiPL-E}}
\newcommand{\HumanevalSynthesize}{\dataset{HumanevalSynthesize}}

\newcommand{\ARCChallenge}{\dataset{ARC-Challenge}}
\newcommand{\HellaSwag}{\dataset{HellaSwag}}
\newcommand{\MMLU}{\dataset{MMLU}}
\newcommand{\TruthfulQA}{\dataset{TruthfulQA}}
\newcommand{\Dolly}{\dataset{Dolly}}
\newcommand{\HHRLHF}{\dataset{HH-RLHF}}

\newcommand{\WinoBias}{\dataset{WinoBias}}
\newcommand{\BOLD}{\dataset{BOLD}}

\definecolor{optimalgreen}{HTML}{2b8a3e}
\usepackage{inconsolata}

\usepackage{amsmath,amsfonts,bm}

\def\eqref#1{equation~\ref{#1}}

\def\1{\bm{1}}

\def\vtheta{{\bm{\theta}}}

\DeclareMathAlphabet{\mathsfit}{\encodingdefault}{\sfdefault}{m}{sl}
\SetMathAlphabet{\mathsfit}{bold}{\encodingdefault}{\sfdefault}{bx}{n}

\title{SCAR: Data Selection via Style Consistency-Aware Response Ranking for Efficient Instruction-Tuning of Large Language Models}

\author{
Zhuang Li$^{1}$, Yuncheng Hua$^{2}$, Thuy-Trang Vu$^{3}$, \\
\textbf{Haolan Zhan$^{3}$, Lizhen Qu$^{3}$, Gholamreza Haffari$^{3}$} \\
$^{1}$School of Computing Technologies, RMIT University, Australia \\
$^{2}$School of Computer Science and Engineering, University of New South Wales \\
$^{3}$Department of Data Science \& AI, Monash University, Australia \\
$^{1}$\texttt{zhuang.li@rmit.edu.au}, $^{2}$\texttt{devin.hua@unsw.edu.au} \\
$^{3}$\texttt{\{trang.vu1, first.last\}@monash.edu}
}

\begin{document}
\maketitle
\begin{abstract}
Recent studies emphasize that manually ensuring a consistent response style and maintaining high data quality in training sets can significantly improve the performance of fine-tuned Large Language Models (LLMs) while reducing the number of training examples needed. However, the precise definition of style and the relationship between style, data quality, and LLM performance remains unclear. This research identifies two key stylistic elements in responses: linguistic form and instructional surprisal. We find that, among training data of comparable quality, higher consistency in these response elements leads to better LLM performance. Inspired by this, we introduce Style Consistency-Aware Response Ranking (\SCAR{}), which automatically prioritizes instruction-response pairs in the training set based on their response stylistic consistency. By selecting the most style-consistent examples, using only 0.7\% of the full dataset in the best case, the fine-tuned LLMs can match or even surpass the performance of models trained on the entire dataset in coding and open-ended question-answering benchmarks. Code and data are available at~\url{https://github.com/zhuang-li/SCAR}. 
\end{abstract}
\section{Introduction}
Instruction-following Large Language Models (LLMs), such as GPT-3.5 and GPT-4~\citep{achiam2023gpt}, have demonstrated strong generalization across diverse language tasks~\citep{chung2022scaling,ouyang2022training}. These models are trained in stages: unsupervised pre-training on large text corpora, followed by supervised fine-tuning (SFT) on instruction-response pairs and additional optimization stages~\citep{bai2022training}.

Recent studies, such as \AlpaGasus~\citep{chen2023alpagasus} and \LIMA~\citep{zhou2023lima}, demonstrate that carefully curated, smaller datasets can outperform larger ones in improving LLM SFT performance. \AlpaGasus finds that smaller datasets with higher quality scores, rated by GPT-4 for helpfulness or correctness, outperform significantly larger ones when used to fine-tune high-capacity LLMs. \textbf{Superficial Alignment Hypothesis}, proposed in \LIMA, suggests that \uline{pre-trained language models already possess the necessary knowledge, and fine-tuning is to guide the model toward specific response styles. Consequently, even a relatively small set of fine-tuning examples might be sufficient to achieve effective alignment.} LIMA demonstrates strong performance by fine-tuning LLMs on just 1,000 high-quality instruction-response pairs, where human experts ensure \textit{stylistic consistency} across responses. However, this hypothesis raises three open questions: (i)~\textit{What key elements constitute response styles that impact LLM SFT?} (ii)~\textit{How does data quality (i.e., helpfulness, correctness) relate to style consistency in influencing efficient SFT?} (iii)~\textit{Can we develop an automatic method that measures stylistic elements to curate smaller, stylistically consistent datasets for more efficient and effective SFT at a lower cost, without relying on human experts?}





Text style is shaped by \textbf{consistent choices} across various linguistic elements~\citep{kang-hovy-2021-style,karlgren2004wheres}, such as lexical, syntactic, and semantic features~\citep{dimarco1993computational}. Our empirical studies have identified two key stylistic factors within responses that significantly affect LLM SFT performance: \textbf{Linguistic Form} and \textbf{Instructional Surprisal}.
\textbf{Linguistic Form} comprises the lexical and syntactic choices that define how a response is presented, independent of its meaning. Empirically, this includes transitional and functional word usage, sentence structure, punctuation patterns, layout features (e.g., headers, bullet points), etc.
\textbf{Instructional Surprisal}, extending from text surprisal measurement~\cite{10.1162/tacl_a_00548,liu-etal-2024-temperature}, in our definition, measures how surprising a response is given an instruction.
We demonstrate that \textit{among SFT datasets with responses at similar levels of helpfulness and correctness, those whose responses exhibit greater consistency in linguistic form and instructional surprisal lead to superior LLM fine-tuning performance.}

Achieving style consistency is challenging, even for human experts. 
We found that datasets containing LLM-generated responses with consistent styles can significantly outperform human-crowdsourced data in enhancing LLM performance.
Therefore, we introduce \textbf{S}tyle \textbf{C}onsistency-\textbf{A}ware Response \textbf{R}anking (\SCAR{}), 
a novel ranking-based model that prioritizes instruction-response pairs with high stylistic consistency and superior data quality.
\SCAR{} is trained on LLM-synthesized and human-crowdsourced datasets to reward responses with higher style consistency regarding linguistic form and instructional surprisal. Enhanced with representation learning, \SCAR{} can better distinguish between these two elements and prioritize aspects that improve LLM performance.  Experiments show that by selecting the most style-consistent examples, using just 0.7\% of the original dataset in some cases, fine-tuned LLMs can match or surpass the performance of models trained on full datasets like \octocoder~\citep{muennighoff2023octopack} and \olmosft~\citep{groeneveld2024olmo} on coding (\HumanEval{}; \citealt{chen2021codex}) and open-ended question answering (\AlpacaEval{}; \citealt{dubois2023alpacafarm}) benchmarks.  

In summary, our key contributions are:

\begin{enumerate}[label=\textsc{\Roman*})]
    \item We identify linguistic form and instructional surprisal as critical response style elements, and demonstrate that within training datasets with comparable helpfulness and accuracy, responses exhibiting higher consistency in linguistic form and instructional surprisal yield better LLMs.
    \item We develop \SCAR{}, a ranking method that selects high-quality, stylistically consistent examples from style-inconsistent datasets. When selecting training data for efficient SFT, \SCAR{} \textbf{outperforms leading data selection baselines}, enabling LLMs trained on small subsets (0.7--25\% of original data) to \textbf{match or exceed full-dataset performance}.
\end{enumerate}
\section{Impact of Styles on LLM Fine-tuning}
\label{sec:style_analysis}
In this section, we study two research questions: i) What key elements in response style can influence LLM SFT? and ii) How do style consistency and data quality impact LLM performance?
\subsection*{RQ1: What Factors Constitute Response Style}
Through empirical analysis of stylistic differences between synthetically generated and human-written instruction-tuning data, we identified two key sets of stylistic features in responses that significantly influence LLM alignment performance.


\begin{itemize}[leftmargin=*]
\item \textbf{\textit{Linguistic Form}} refers to the structure of language, including how words and sentences are organized and interact~\citep{FABB20018292,Chomsky1957,jurafsky2000speech}. In our context, it denotes elements that shape the presentation of a response, mostly independent of semantics, such as transitional and functional word usage, tone, sentence structure, punctuation patterns, layout features (e.g., headers, bullet points), etc. For example, we observe that \gptThreeFive responses often follow a consistent structure, using bullet points and similar transitional phrases, whereas human responses, authored by diverse individuals, show greater variation in linguistic elements.

\item \textbf{\textit{Instructional Surprisal}} measures how surprising the content (solutions, ideas, and approaches) of a response is in addressing a given instruction. For example, when asked about sorting algorithms, \gptThreeFive consistently provides predictable solutions like merge sort or quick sort, while human responses show a range of surprisal--from conventional approaches to unexpected choices like StoogeSort or novel answers. Instructional surprisal can be estimated using perplexity scores, computed as the average negative log-likelihood of the response given the instruction, or through semantic relatedness metrics such as cosine similarity between instruction and response embeddings. These approaches extend word-level surprisal measures based on language model predictions~\cite{10.1162/tacl_a_00548,liu-etal-2024-temperature} or word-to-context embedding similarity~\cite{sayeed2015vector,karampiperis2014towards} to the sequence level. Further discussion is provided in Appendix~\ref{app:surprisal_modelling}.
\end{itemize}

\subsection*{RQ2: Influence of Style Consistency and Data Quality on LLM Performance}
\label{sec:RQ2}
We collect both human-written and synthetic data in coding and general open-ended domains, and conduct stylometric and quality analyses on this data. Following this, we fine-tune base LLMs using this data to explore how style consistency and data quality influence LLM SFT performance.

We control style variations to create three dataset types--\textcolor{blue}{human-written}, \textcolor{red}{referenced}, and \textcolor{green}{direct}--to explore how linguistic form and response surprisal impact LLM performance. In the coding domain, we collect 10,000 human-written instruction-response pairs from \StackExchange{}\footnote{https://stackexchange.com/}, an online platform that includes 11 million pairs of coding questions and answers. We use the \LIMA{} dataset, including 1,000 human-generated examples, for the general domain. Additionally, we generate two synthetic response types with controlled styles: ``referenced'' and ``direct.'' ``Referenced'' responses are crafted by a chat-LLM that rewrites human responses to retain their semantic meaning, similar to the method in~\citet{yang-etal-2024-self}. This process retains the surprisal levels of human responses but alters their linguistic form. We also filter out examples where ``referenced'' responses deviate significantly from human responses in surprisal metrics, reducing the datasets to 944 \StackExchange{} and 407 \LIMA{} examples. In contrast, the chat-LLM generates ``direct'' responses to the remaining instructions without any references, potentially producing different semantics, thereby significantly varying their surprisal levels compared to human-referenced responses.
Table~\ref{tab:fifa_instant_replay_dense} illustrates the similarities and differences among these three style variants.
\begin{table}[h]
\centering
\renewcommand{\arraystretch}{1}
\scriptsize
\begin{tabular}{|p{0.92\columnwidth}|}
\hline
\textbf{Instruction} \\
\hline
Why is FIFA against adding instant replay to the game? \\
\hline
\textbf{Human Response} \\
\hline
There are a few fundamental considerations:

\textcolor{orange}{Uncertainty is a part of the game.} ...
\textcolor{orange}{Football is global.} ...
\textcolor{orange}{Football is a fast-paced game with few opportunities for stoppage.} ...
In 2008, FiFA President Sepp Blatter said: 
...
\\
\hline
\textbf{Human Referenced Response from GPT-3.5-turbo} \\
\hline
\textcolor{optimalgreen}{FIFA is against adding instant replay to the game for several reasons}. One of the main concerns is that \textcolor{orange}{uncertainty is seen as a natural part of the game},... \textcolor{optimalgreen}{Additionally}, \textcolor{orange}{football is a global sport};... Another factor is that \textcolor{orange}{football is a fast-paced game with minimal stoppages};... \\
\hline
\textbf{Direct Response from GPT-3.5-turbo} \\
\hline
\textcolor{optimalgreen}{FIFA is against adding instant replay to the game} because they believe it would disrupt the flow of the game ... They also argue that human error is a part of the game ... \textcolor{optimalgreen}{Additionally}, implementing instant replay would require significant changes to the rules and regulations ... \\
\hline
\end{tabular}
\vspace{-2mm}
\caption{Examples of different response types for a given instruction. Some details are abbreviated as `...'. Shared surprisal-related style elements between ``Human'' and ``Referenced'' responses are highlighted in \textcolor{orange}{orange}, and shared linguistic form elements between ``Referenced'' and ``Direct'' responses are in \textcolor{optimalgreen}{green}.}

\label{tab:fifa_instant_replay_dense}
\end{table}



 We also isolate the effects of data quality on LLM performance by using three chat-LLMs with different capabilities to generate synthetic ``referenced'' and ``direct'' datasets. The models employed are \gptThreeFive, \llamaTwoSeventyB, and \llamaTwoThirteenB~\citep{touvron2023llama}, with \gptThreeFive being the most advanced, followed by \llamaTwoSeventyB and \llamaTwoThirteenB, according to the arena-leaderboard~\citep{zheng2024judging}. We find that hallucinations occurring in LLM-generated ``referenced'' and ``direct'' responses can significantly affect the quality of the resulting synthetic data.

\paragraph{Stylometric Analysis.}
\textit{To analyze the linguistic form of human and synthetic responses}, we employ six authorship attribution metrics~\citep{tripto-etal-2023-hansen,zheng2023review} that capture non-semantic features. These include the Type Token Ratio (TTR)~\citep{templin1957certain}, Measure of Textual Lexical Diversity (MTLD)~\citep{mccarthy2005assessment} for functional words, Flesch score~\citep{kincaid1975derivation}, average sentence length, and the frequency of punctuation and layout features (e.g., bullet points and headers). Higher TTR and MTLD values indicate greater lexical diversity, while a higher Flesch score suggests improved readability. We identify functional words in the response using a lexicon based on heuristic POS-tagging rules. \textit{To assess instructional surprisal}, we compute the perplexity of a response given its instruction, denoted as $\text{PPL}(y \mid x)$, using \llamaThreeEightB~\citep{llama3modelcard}. 


\begin{figure}
\vspace{-2mm}
    \centering
    \includegraphics[width=\linewidth]{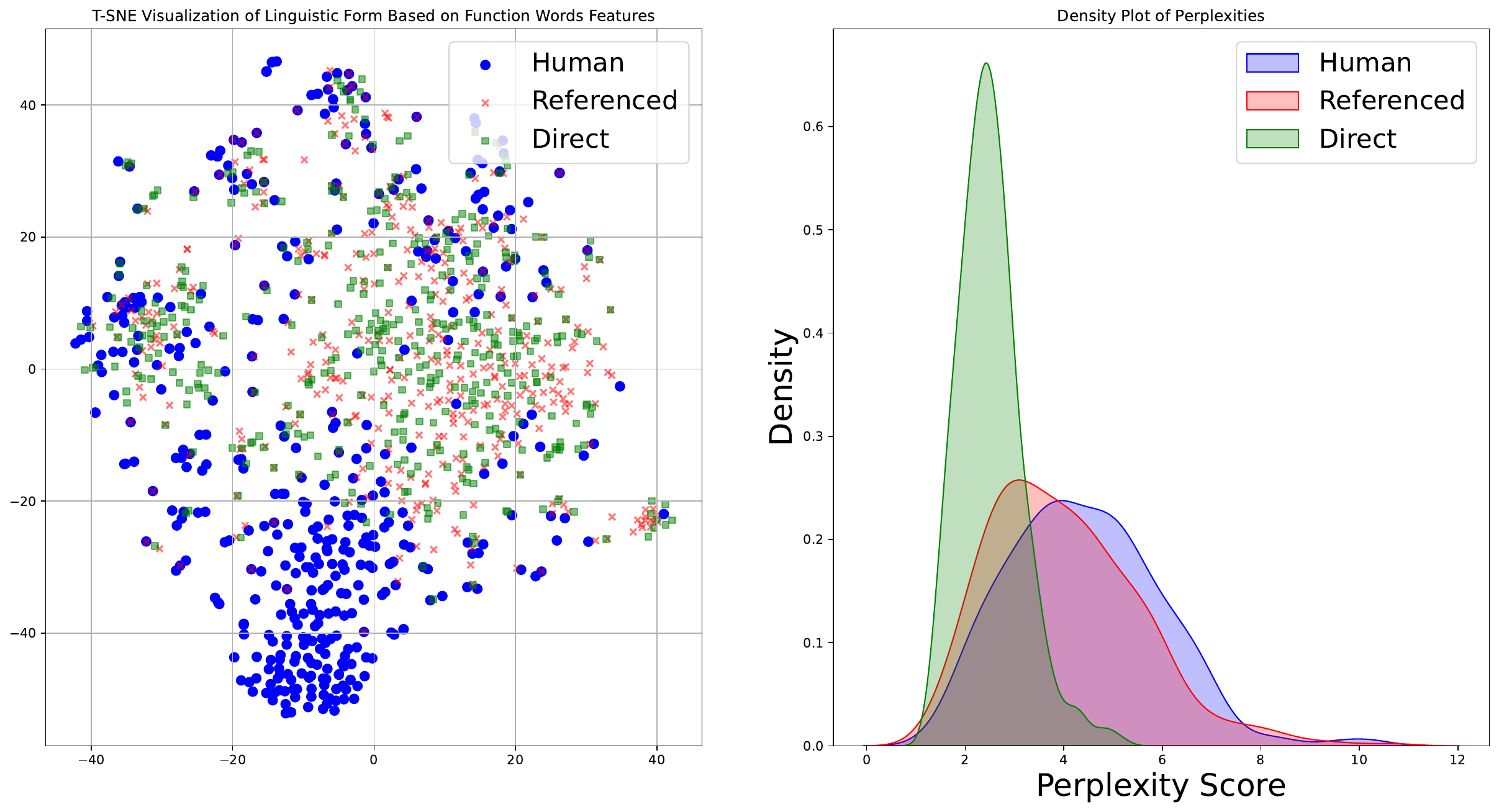}
   \vspace{-3mm}
    \caption{(Left) T-SNE plot showing embeddings of the linguistic forms of human and \textsc{GPT-3.5-turbo} responses to LIMA instructions. (Right) Density plot of perplexity detailing the surprisal levels of the responses.}
    \label{fig:presentation_styles}
 \vspace{-7mm}
\end{figure}
\begin{table*}[t!]
\centering
\resizebox{\textwidth}{!}{
\begin{tabular}{l|ccc|ccc}
\toprule
 & \multicolumn{3}{c|}{\textbf{\StackExchange}} & \multicolumn{3}{c}{\textbf{\LIMA}}  \\
\cline{2-7}
  \multirow{3}{*}{\begin{tabular}[c]{@{}l@{}}\textbf{Data Curation} \\ \textbf{Methods}\end{tabular}} & 
  \multicolumn{1}{c}{\begin{tabular}[c]{@{}c@{}}\textbf{Stylometric} \\ \textbf{Analysis}\end{tabular}} &
  \multicolumn{1}{c}{\begin{tabular}[c]{@{}c@{}}\textbf{Data} \\ \textbf{Quality}\end{tabular}} & 
  \multicolumn{1}{c|}{\begin{tabular}[c]{@{}c@{}}\textbf{\codeLlamaSevenB} \\ \textbf{Performance}\end{tabular}} & 
  \multicolumn{1}{c}{\begin{tabular}[c]{@{}c@{}}\textbf{Stylometric} \\ \textbf{Analysis}\end{tabular}} & 
  \multicolumn{1}{c}{\begin{tabular}[c]{@{}c@{}}\textbf{Data} \\ \textbf{Quality}\end{tabular}} & 
  \multicolumn{1}{c}{\begin{tabular}[c]{@{}c@{}}\textbf{\llamaThreeEightB} \\ \textbf{Performance}\end{tabular}} \\
\cdashline{2-7}
 & \multicolumn{1}{c}{\begin{tabular}[c]{@{}c@{}}\textbf{Std. TTR} $\downarrow$ /\\ \textbf{Std. PPL($y|x$)}$\downarrow$\end{tabular}}  & 
 \multicolumn{1}{c}{\begin{tabular}[c]{@{}c@{}}\textbf{Helpfulness} /\\ \textbf{Correctness}\end{tabular}} & 
 \multicolumn{1}{c|}{\begin{tabular}[c]{@{}c@{}}\textbf{Avg. Pass@1} /\\ \textbf{Avg. Pass@10}\end{tabular}} & 
 \multicolumn{1}{c}{\begin{tabular}[c]{@{}c@{}}\textbf{Std. TTR}$\downarrow$ /\\ \textbf{Std. PPL($y|x$)}$\downarrow$\end{tabular}} & 
 \multicolumn{1}{c}{\begin{tabular}[c]{@{}c@{}}\textbf{Helpfulness} /\\ \textbf{Correctness}\end{tabular}} & 
 \textbf{L.C. WinRate} \\
\hline
\hline
Human Response & 24.23 / 0.33 & 3.29 / 3.70 & 26.56 / 41.63 & 20.49 / 1.53 & 3.86 / 4.14 & 1.93 \\
\hline
\multicolumn{1}{l|}{\gptThreeFive} & \multicolumn{3}{c|}{} & \multicolumn{3}{c}{}  \\ 

\ \ \ \ Referenced & 8.16 / 0.33 & 3.44 / 3.70 & 29.82 / 46.89 & 18.43 / 1.52 & 3.79 / 4.00 & 3.64 \\

\ \ \ \ Direct & 8.14 / 0.30 &  3.32 / 3.45 & 31.00 / 47.12 & 16.06 / 0.64 & 3.91 / 4.16 & 5.67 \\
\hline
\multicolumn{1}{l|}{\llamaTwoSeventyB} & \multicolumn{3}{c|}{} & \multicolumn{3}{c}{}\\ 

\ \ \ \ Referenced & 11.90 / 0.36  & 3.14 / 3.54  & 29.82 / 44.03 & 16.51 / 1.45 & 3.89 / 4.11 & 3.96 \\

\ \ \ \ Direct & 13.52 / 0.28  & 3.18 / 2.71  & 30.89 / 45.31 & 15.63 / 0.42 & 3.85 / 4.22 & 6.25 \\
\hline
\multicolumn{1}{l|}{\llamaTwoThirteenB} & \multicolumn{3}{c|}{} & \multicolumn{3}{c}{}\\ 

\ \ \ \ Referenced & 7.46 / 0.27  & 2.65 / 2.68  & 26.61 / 41.91 & 13.64 / 1.19 & 3.75 / 3.89 & 3.77 \\

\ \ \ \ Direct & 8.86 / 0.28  & 1.85 / 1.70  & 26.42 / 40.00 & 14.22 / 0.38 & 3.29 / 3.48 & 6.22 \\
\bottomrule
\end{tabular}
}
\vspace{-3mm}
\caption{Performance comparison of \codeLlamaSevenB and \llamaThreeEightB fine-tuned on \StackExchange{} and \LIMA{} instructions paired with human responses and two variants (`Referenced'/`Direct') generated by three chat-LLMs, analyzing response quality (helpfulness/correctness) and style consistency (TTR/PPL standard deviations).}
\label{tab:one_column_performance_comparison}
\vspace{-4mm}
\end{table*}

T-SNE~\citep{van2008visualizing} plots (Figure~\ref{fig:presentation_styles}, left) show that embeddings of \gptThreeFive-generated \textcolor{red}{``referenced''} and \textcolor{optimalgreen}{``direct''} responses cluster tightly in the center, indicating that both synthetic response types share consistent and similar \textit{linguistic forms}. These embeddings are created by vectorizing six authorship metrics and the unigrams of functional words. Conversely, \textcolor{blue}{human} responses are more dispersed in the outer region, showing lower consistency. Figure~\ref{fig:presentation_styles} (right) shows \textcolor{optimalgreen}{``direct''} responses have a more skewed perplexity distribution towards lower values, indicating higher consistency in \textit{instructional surprisal} compared to both \textcolor{red}{``referenced''} and \textcolor{blue}{human} ones. 

Standard deviations (Std.) of TTR and perplexity for different response types are listed in Table~\ref{tab:one_column_performance_comparison}, with additional linguistic form and text surprisal metrics detailed in Table~\ref{tab:style_analysis_expand} (\Cref{appx:full-style-analysis}). We observe human responses have higher Std. values regarding TTR, perplexity and other metrics compared to synthetic responses, and ``referenced'' responses show a higher perplexity Std. than ``direct'' responses. The Std. values of these metrics across ``referenced'' and ``direct'' responses from \llamaTwoSeventyB, \llamaTwoThirteenB, and \gptThreeFive indicate synthetic responses from \textbf{all} these LLMs have higher consistency in both stylistic elements than human ones.



\paragraph{Data Quality Analysis.} 
We evaluate a sample of 100 examples from each dataset using \textsc{gpt-4-1106-preview}. We rate the scores for two data quality metrics, \textit{helpfulness} and \textit{correctness}, using the adjusted prompt from the automatic data evaluator ICE-Score~\citep{zhuo2024ice} for the coding domain and AlpaGasus~\citep{chen2023alpagasus} for the open-ended domain, and then calculate the average scores across the samples. Higher scores indicate better quality. 
Table~\ref{tab:one_column_performance_comparison} reveals that in the coding domain, \gptThreeFive-generated responses match the quality of human-written ones, while other LLMs produce lower-quality data. 
In the open domain, \llamaTwoSeventyB and \gptThreeFive responses are comparable in quality to human-written responses, whereas \llamaTwoThirteenB responses are of slightly lower quality.

\paragraph{Impact on LLM Performance.}
We evaluate the \codeLlamaSevenB model fine-tuned with LoRA~\citep{hu2021lora} on various datasets using \HumanEval{} (Python) \citep{chen2021codex} and \MultiPLE{} (Java, JavaScript, C++) \citep{cassano2023multipl} benchmarks. For the coding domain, we report average Pass@1 and average Pass@10 execution accuracies across all coding questions spanning four programming languages. We measure the length control win rate (L.C. WinRate) \citep{dubois2024length} by comparing responses from the LoRA fine-tuned \llamaThreeEightB with those from \textsc{GPT-4-preview-1106} on 2500 open-domain instructions from AlpacaEval\footnote{https://github.com/tatsu-lab/alpaca\_eval/}. We use \llamaThreeSeventyB~\citep{llama3modelcard} as our automatic evaluator for its cost-effectiveness (\$0.9 per evaluation). This evaluator correlates with human judgment as well as GPT-4 evaluators and even surpasses human-to-human agreement, with an agreement rate of 67.5\% compared to 65.7\% in tests conducted on \AlpacaEval{}.



When comparing synthetic responses of similar or slightly different quality generated from capable chat-LLMs, ``direct'' responses outperform their ``referenced'' counterparts in downstream LLM SFT tasks through higher instructional surprisal consistency. 
Both synthetic types exhibit greater consistency in both stylistic elements, thereby outperforming human-authored data.
However, style consistency alone cannot compensate for substantial quality deficits. This is evidenced by a notable exception in coding tasks, where \llamaTwoThirteenB's ``direct'' responses, despite having higher style consistency, achieve poorer fine-tuning outcomes due to their significantly lower quality scores (1.8) compared to both ``referenced'' responses (2.6) and human data (3.5).


\paragraph{Takeaway.}
The analysis reveals several insights:
\begin{enumerate}[label=\textsc{\Roman*})]
    \item \textbf{\textit{Linguistic form}} and \textbf{\textit{instructional surprisal}} inherent in the response styles of the training data significantly influence the LLM SFT performance.
    \item LLM-generated responses show higher style consistency than human ones, with ``direct'' responses showing the greatest consistency in \textbf{\textit{linguistic form and instructional surprisal}}.
\item Enhancing data quality and ensuring response style consistency both contribute to improved LLM SFT performance. Among datasets with shared instructions and similar quality, those with more consistent response styles yield better LLM performance.

\end{enumerate}



\section{Style Consistency-Aware Ranking}
Inspired by these findings, we develop a Style Consistency-Aware Ranker to select training examples with consistent response styles, improving LLM SFT performance.

\paragraph{Ranking Objective.} Given a dataset $\mathcal{D} = \{(x_i, y_i^{d}, y_i^{r}, y_i^{h})\}_{i=1}^N$, where $x_i$ represents the instruction, $y_i^{d}$ and $y_i^{r}$ are the ``direct'' and human ``referenced'' responses from chat-LLMs, respectively, and $y_i^{h}$ represents the human response. \textbf{We aim to learn a ranking function \( R(x, y) \) that assigns higher scores to high-quality responses adhering to the consistent style of a specific LLM's outputs.} The objective for each instance is to learn the ranking function: 

\begingroup
\small
\begin{align}
\nonumber
\mathcal{L}_{r}(x, y^d, y^r, y^h) & = \\
\sum_{(y^a, y^b) \in \mathcal{P}}  \max(0, \alpha & - R_{\vtheta}(x, y^a) + R_{\vtheta}(x, y^b)) \\
\text{s.t.} 
\quad \min (f(x, y^a) & ,\ f(x, y^b)) > \sigma 
\label{eq:constraint}
\end{align}
\endgroup
\noindent where $\mathcal{P} = \{(y^d, y^r), (y^r, y^h), (y^d, y^h)\}$ represents the set of desired pairwise orderings, based on the findings from Section~\ref{sec:style_analysis}, that ``direct'' responses are more consistent in surprisal levels than ``referenced'' ones, ``referenced'' responses are more consistent in linguistic form than human data, and ``direct'' responses are more consistent than human data in both stylistic feature types. The margin $\alpha$ ensures the difference in the ranking scores assigned by $R_{\vtheta}(x, y)$, while the quality measure function $f(x, y)$ evaluates the quality (e.g., helpfulness, correctness) of the response $y$ given the instruction $x$. The quality measure function $f$ can be implemented using strong LLMs such as \textsc{GPT-3.5} or \textsc{GPT-4} with a prompt, as in \citet{chen2023alpagasus}, to evaluate the helpfulness and correctness of the answers and average these scores to obtain the final quality score. The quality threshold $\sigma$ ensures the ranker only rewards responses that are \textbf{both style-consistent and high-quality}. 

\paragraph{Reward Function.} The reward function \( R_{\vtheta}(x, y) \) is modelled as a neural network that takes representations of instructional surprisal \( \mathbf{v}_c \in \mathbb{R}^{1 \times M} \) and linguistic form \( \mathbf{v}_p \in \mathbb{R}^{1 \times M} \), and computes a scalar reward score using a multi-layer perceptron (MLP):
\begin{align}
\nonumber
R_{\vtheta}(x, y) &= \text{MLP}_r([\mathbf{v}_p; \mathbf{v}_c]) \\
\nonumber
\mathbf{v}_p &= \text{Max-Pool}(\mathbf{V}_y) \\
\mathbf{v}_c &= \text{MLP}_c([\mathbf{V}^{0}_x; \mathbf{V}^{0}_y])
\label{eq:ranker_model}
\end{align}
Our experiments demonstrate that linguistic form, when compared to semantic content, shows both minimal influence on the variance of instructional surprisal and significantly lower instruction dependence. These findings motivate us to adopt disentangled modeling strategies. For linguistic form, we capture surface-level features through max pooling over the response sequence \( \mathbf{V}_y \), independent of the instruction. For instructional surprisal, we approximate related features using semantic relatedness, motivated by prior work that models surprisal through embedding similarities between words and their contextual text~\citep{sayeed2015vector}. Specifically, we compute multi-dimensional semantic alignment by concatenating the \texttt{[CLS]} embeddings of the instruction \( \mathbf{V}^{0}_x \) and response \( \mathbf{V}^{0}_y \), and processing them through $\text{MLP}_c$. To generate sequence representations \( \mathbf{V} \), we use a pre-trained encoder, such as \textsc{RoBERTa-base}~\citep{liu2019roberta} and \textsc{CodeT5p-110M-Embedding}~\citep{wang2023codet5+}. Further details are provided in Appendix~\ref{app:cmi_calculation} and~\ref{app:surprisal_modelling}.



\paragraph{Style Representation Learning.} 
Accurately capturing distinct representations for linguistic form (\( \mathbf{v}_{p} \)) and instructional surprisal (\( \mathbf{v}_{c} \)) is challenging, as these features can still become entangled during the learning process, even with our specialized separation design. To address this, we leverage observed similarities: the linguistic form of ``referenced'' responses is more similar to ``direct'' responses than to human responses, and the instructional surprisal of ``referenced'' responses is closer to that of human responses than to ``direct'' ones, as shown in Figure~\ref{fig:presentation_styles}. We introduce a regularization term using triplet margin losses to enforce these similarity patterns:
\begin{align}
\nonumber
\mathcal{L}_{rl}(x, y^d, y^r, y^h) &= \\
\nonumber
\lambda_p \max \{0, d(\mathbf{v}_{p}^{d}, & \mathbf{v}_{p}^r) - d(\mathbf{v}_{p}^r, \mathbf{v}_{p}^h) + \beta_{p}\} \\
+ \lambda_c \max \{0, d(\mathbf{v}_{c}^{h}, & \mathbf{v}_{c}^r) - d(\mathbf{v}_{c}^d, \mathbf{v}_{c}^h) + \beta_{c}\}
\label{eq:disentanglement}
\end{align}
\noindent where \( d(\mathbf{v}_i, \mathbf{v}_j) = \left\|\mathbf{v}_i - \mathbf{v}_j\right\|_2 \) is the distance function and $\beta$ values are the margins.

\paragraph{Final Loss Function.} The final loss function combines the ranking loss and the representation learning losses:
$\mathcal{L}_{scar} = \mathcal{L}_{r} + \mathcal{L}_{rl}$ 

\paragraph{Ranking and Filtering.} After training reward function $R_{\vtheta}(x, y)$, it ranks instruction-response pairs $(x, y)$ in a held-out dataset. The top $k\%$ of examples with the highest scores are selected to create a high-quality style-consistent subset for fine-tuning LLMs. This filtered dataset is expected to \textbf{\textit{improve fine-tuned LLM performance on target tasks to levels comparable to or exceeding those achieved using the entire original dataset}}. 
\section{Experiments}

We train \SCAR{} using data from the \textit{\textbf{coding}} and \textit{\textbf{open-ended question-answering}} domains to select examples for LLM SFT from the full dataset in these same domains.

\paragraph{Ranker Data.}
We collect instructions for \SCAR{} training and evaluation, which include 10,000 randomly selected examples from \StackExchange{} for the code domain, and 6,000 instructions from a combination of 5,000 random \Dolly{}~\citep{DatabricksBlog2023DollyV2} data samples and the full LIMA dataset. \Dolly{} is a human-curated dataset with 15,000 high-quality instruction-response pairs. We create the data by pairing instructions with human responses and the ``referenced'' and ``direct'' responses generated by \gptThreeFive, as described in Section~\ref{sec:style_analysis}. Due to budget limitations, we use \gptThreeFive to rate the helpfulness and correctness of responses according to the constraint in Eq.~(\ref{eq:constraint}). 


\paragraph{LLM SFT Data.}
\SCAR{} and other baselines select data from two sources, held out from the ranking training data. These sources provide diverse but style-inconsistent examples: 
\textit{\textbf{i) Human-Crowdsourced Data}}, curated by many authors, making it diversified and naturally style-inconsistent. 
\textit{\textbf{ii) Mixed Synthetic Data}}, generated by \gptThreeFive using various system prompts, reflecting the practical use of multiple open-source synthetic datasets to enhance diversity. 

\textit{\textbf{For the code domain}}, human-written data comes from a sample of 20,000 crowdsourced \StackExchange{} examples. To ensure quality, we select examples with instructions that include code blocks and answers with a rating above 2. 

The mixed synthetic data comprises 20,000 examples, sourced evenly from: 
\textbf{i)} 5,000 \StackExchange{} instructions with ``direct'' responses,
\textbf{ii)} 5,000 \StackExchange{} instructions with ``referenced'' responses, 
\textbf{iii)} 5,000 coding examples curated using \textsc{Evol-Instruct}~\citep{luo2023wizardcoder} by~\citet{zan2023codem}, 
\textbf{iv)} 5,000 coding examples generated using \textsc{Self-Instruct}~\citep{wang2023self}. 

The instructions cover Python, Java, JavaScript, and C++. For \textsc{Self-Instruct}, we use \gptThreeFive to generate responses in the target programming languages identified using \texttt{guesslang}\footnote{\url{https://github.com/yoeo/guesslang}}.

\textit{\textbf{For the open-ended domain}}, human-written data comes from 10,000 \Dolly{} examples, held out from the \Dolly{} examples used for ranker training. 

Mixed synthetic data includes 10,000 examples, evenly sourced from: 
\textbf{i)} 2,500 held-out \Dolly instructions with ``direct'' answers, 
\textbf{ii)} 2,500 \Dolly instructions with ``referenced'' answers, 
\textbf{iii)} 2500 open-domain examples using \textsc{Self-Instruct} by LaMini~\citep{wu2023lamini}, 
\textbf{iv)} examples curated using \textsc{Evol-Instruct} from~\citet{xu2023wizardlm}.

\paragraph{Data Selection and LLM SFT.} The data selection methods sample 50\%, 25\%, and 12.5\% of coding-domain data to fine-tune \codeLlamaSevenB, and 50\%, 25\%, and 10\% of open-domain data to fine-tune \llamaThreeEightB. Both LLM trainings use LoRA due to computational constraints.

\paragraph{LLM Evaluation.} We evaluate code generation performance using \HumanEval and \MultiPLE, and report the metric:
\[
\text{Avg.\ Pass@($1{+}10$)} = \frac{\text{Pass@1} + \text{Pass@10}}{2}
\]
averaged over four languages for the fine-tuned \codeLlamaSevenB. For general tasks, we use \AlpacaEval and report the L.C. WinRate of outputs from fine-tuned \llamaThreeEightB compared to \textsc{GPT-4-preview-1106}, as in Section~\ref{sec:style_analysis}.

\paragraph{Data Selection Baselines.}
\begin{figure*}[t]
    \centering
    \includegraphics[width=\textwidth]{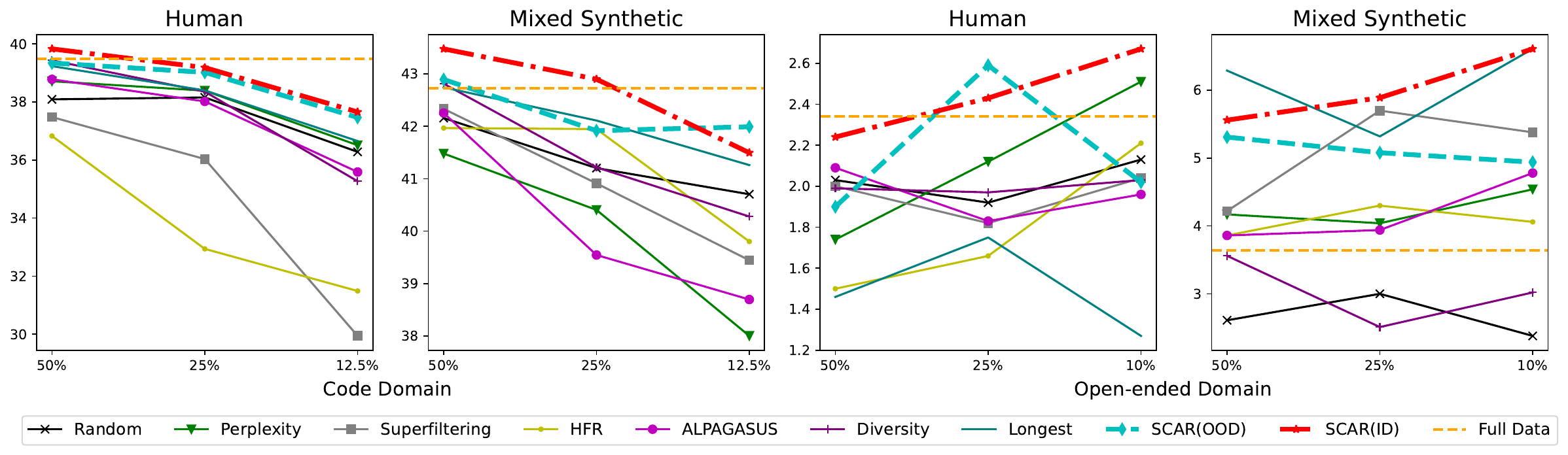}
 \vspace{-7mm}
\caption{The performance of LLMs fine-tuned on human and synthetic data subsets of various sizes in code and open domains, sampled with different data selection approaches.}
    \label{fig:main_res}
 \vspace{-3mm}
\end{figure*}
We compare \SCAR{} in two settings with 7 baselines:
\begin{enumerate}[leftmargin=*]
   \item  \textbf{\Random}: Randomly select examples.
   \item \textbf{\Perplexity}~\citep{albalak2024survey}: Select entries with lowest response perplexity ($\text{PPL}(y|x)$) computed using \llamaThreeEightB.
\item \textbf{\Superfiltering}~\citep{li2024superfiltering}: Select the most challenging examples for LLMs with the highest Instruction-Following Difficulty (IFD) score. Here, we compute IFD as $\frac{\text{PPL}(y|x)}{\text{PPL}(y)}$  using \llamaThreeEightB.
\item \textbf{\textsc{Human Feedback Ranking (HFR)}}: Uses the same ranker architecture as \SCAR{}, trained on 10,000 Stack Exchange pairs~\citep{h4stackexchange} annotated with human preferences (each instruction paired with positive and negative responses) for the coding domain, and 6,000 human preference examples from \HHRLHF data~\citep{bai2022training} for the general domain.
\item \textbf{\AlpaGasus}~\citep{chen2023alpagasus}: Select data based on response quality scores rated by \gptThreeFive, consistent with the rating method used in our ranker.
\item \textbf{\Diversity}: Apply k-means clustering to diversify examples by selecting randomly from each cluster, a method commonly used in active learning~\citep{li2023active,li2023best,zhdanov2019diverse}.
\item \textbf{\Longest}: Select examples with longest response token lengths~\citep{zhaolong}.
\item \textbf{\SCARID}: \SCAR{} trained on in-domain (ID) data (e.g., code) and selects examples within the same domain.
\item \textbf{\SCAROOD}: \SCAR{} trained on in-domain data and select examples from an out-of-domain (OOD) dataset. For instance, \SCAROOD is trained on the code domain and selects data from the open domain or vice versa.
\end{enumerate}

\subsection{Main Results and Discussion}
\paragraph{Effectiveness of \SCAR{}-Selected Data.}
As in Figure~\ref{fig:main_res}, \SCARID can enhance SFT performance while lowering computational costs. LLMs fine-tuned on only 25\% and 10\% of \SCARID-selected data achieve comparable or superior performance to models trained on full datasets in coding and general domains, respectively. 

\SCARID and \SCAROOD generally outperform other data selection methods for fine-tuning LLMs, with \SCAROOD slightly lagging behind \SCARID due to challenges in cross-domain generalization. Some baselines show unstable performance. \Superfiltering performs poorly in the coding domain. We observe it may assign high IFD scores to erroneous examples in crowdsourced coding data of varying quality. \Perplexity and \AlpaGasus-selected data result in similar LLM performance trends. However, their performance is inferior to \SCARID, which we attribute to their lack of style consistency. Traditional active learning methods like \Random and \Diversity sampling prove less effective, as our style-inconsistent target scenario inherently incorporates diversity, limiting their additional benefits. \HFR's underperformance across most scenarios suggests that training the ranker on inconsistent human preferences from diverse authors may impair its ability to select optimal training data. Notably, \Longest performs comparably to our method in open-domain synthetic data selection, though inferior elsewhere. This aligns with our style consistency framework, as length serves as a strong style indicator, with \textsc{Evol-Instruct} responses consistently being longer.


\paragraph{Impact of Data Sizes.} Figure~\ref{fig:main_res} shows that in the coding domain, using fewer data selected by various methods usually lowers LLM performance. However, in the open-ended domain, most methods can select fewer synthetic data to fine-tune LLMs that outperform those trained on the full dataset. With \SCARID, reducing data consistently improves LLM performance in the open domain.  This suggests that while dataset size, diversity, and style consistency can all benefit LLM SFT, their optimal balance varies across different scenarios.

\begin{table}
\centering
\renewcommand{\arraystretch}{0.75}
\resizebox{0.95\columnwidth}{!}{
\begin{tabular}{r|c|c|c|c}
\toprule
& \textbf{Std. TTR} & \textbf{Std. PPL} & \textbf{Helpful} & \textbf{Correct} \\
\hline
\hline
\multicolumn{5}{c}{\textbf{\StackExchange}} \\
\hline
100\% & 21.48 & 1.80 & 2.84 & 2.68 \\
50\% & 16.78 & 1.61 & 3.02 & 3.01 \\
25\% & 14.85 & 1.61 & 2.78 & 2.72 \\
12.5\% & 14.29 & 1.94 & 2.67 & 2.77 \\
\hline
\multicolumn{5}{c}{\textbf{\Dolly}} \\
\hline
100\% & 30.96 & 65.70 & 3.95 & 3.91 \\
 50\% & 28.43 & 54.32 & 3.98 & 3.99 \\
 25\% & 24.74 & 49.51 & 3.96 & 3.93 \\
 10\% & 23.73 & 39.58 & 3.98 & 3.99 \\
\bottomrule
\end{tabular}
}
\vspace{-3mm}
\caption{Stylometric and quality analysis of data subsets selected by \SCARID from the full human-crowdsourced \StackExchange and \Dolly datasets.}
\label{tab:style_analysis_scar}
\end{table}

\paragraph{Stylometric and Data Quality Analysis of SCAR-Selected Data.} Table~\ref{tab:style_analysis_scar} shows that \SCARID improves style consistency in the selected \Dolly data, reflected by consistently lower TTR and perplexity standard deviation compared to the full dataset. However, for code data, while the TTR standard deviation decreases, the perplexity standard deviation increases when selecting smaller subsets (25\%, 12.5\%), suggesting that differentiating instructional surprisal features in code is challenging. This may explain the sudden performance drop in LLMs fine-tuned on these smaller code subsets. Moreover, our method preserves average data quality (helpfulness, correctness), as rated using \textsc{GPT-4-1106-preview}, comparable to the full dataset, likely due to the use of the data quality constraint in Eq.~(\ref{eq:constraint}) during ranker training.


\begin{table}
\centering
\resizebox{0.95\columnwidth}{!}{
\begin{tabular}{c|c|c|ccc}
\toprule
\multirow{2}{*}{\textsc{OLMO-7b}} & Data Sizes & 320k &  10k & 5k & 2.5k  \\
& L.C. WinRate & 3.86 & 5.37 & 5.64  & 4.08   \\
\hline
\hline
\multirow{2}{*}{\textsc{Starcoder-15.5b}} &  Data Sizes    & 13k &  10k & 5k & 2.5k  \\
& Avg. Pass@(1+10) & 37.85  & 39.69 & 40.09  & 40.14  \\
\bottomrule
\end{tabular}
}
\vspace{-3mm}
\caption{L.C. WinRate for \olmo and Avg. Pass@(1+10) for \starcoder fine-tuned on original (320k, 13k) and subset sizes (10k, 5k, 2.5k).}
\vspace{-3mm}
\label{tab:open_llm_comparing}
\end{table}
\paragraph{Effectiveness of \SCAR{} on Open-Source LLMs.}
Specifically, we apply the \SCARID method to select 2.5k, 5k, and 10k instruction–response pairs from the \texttt{allenai/tulu-v2-sft-mixture} (320k examples) and \texttt{bigcode/guanaco-commits} (13k examples), after removing non-English entries and exact duplicates. Both datasets contain a high degree of stylistic inconsistency in the responses, due to either merging multiple existing datasets (Tulu) or scraping content authored by many individuals (Guanaco-commits). We then compare their performance to the official checkpoints, \olmosft and \octocoder~\citep{muennighoff2023octopack}, which were instruction-tuned on the full datasets.
Table~\ref{tab:open_llm_comparing} shows that \SCAR{}-selected subsets significantly boost performance, achieving these results with only 0.7\% to 20\% of the original data, as measured by L.C. WinRate on \AlpacaEval and average Pass@(1+10) on \HumanEval and \MultiPLE. Further evaluation of \olmo variants on diverse benchmarks (Table~\ref{tab:benchmark_comparison_four}, Appendix~\ref{app:four_bench})--including \ARCChallenge~\cite{clark2018think}, \TruthfulQA~\cite{lin2022truthfulqa}, \HellaSwag~\cite{zellers2019hellaswag} and \MMLU~\cite{hendrycksmeasuring}--demonstrates that \textbf{all} our subset-fine-tuned \olmo outperform the full 320k-trained model in average performance across various LLM capabilities. 
\subsection{Ablation Study}
To evaluate the effectiveness of \SCARID components, we compare the full ranker training setting (Full, GPT-3.5) against variations without the quality constraint in Eq.~(\ref{eq:constraint}) (w/o con, GPT-3.5), without representation learning in Eq.~(\ref{eq:disentanglement}) (w/o rl, GPT-3.5), and without ``referenced'' responses during training (w/o ref, GPT-3.5). We also generate synthetic data to train the ranker using \llamaTwoThirteenB (Full, Llama2-13b), \llamaTwoSeventyB (Full, Llama2-70b), \llamaThreeSeventyB (Full, Llama3-70b), and \llamaTwoThirteenB without using quality constraint (w/o con, Llama2-13b).

\begin{figure}[t]
    \centering
    \includegraphics[width=1.0\columnwidth]{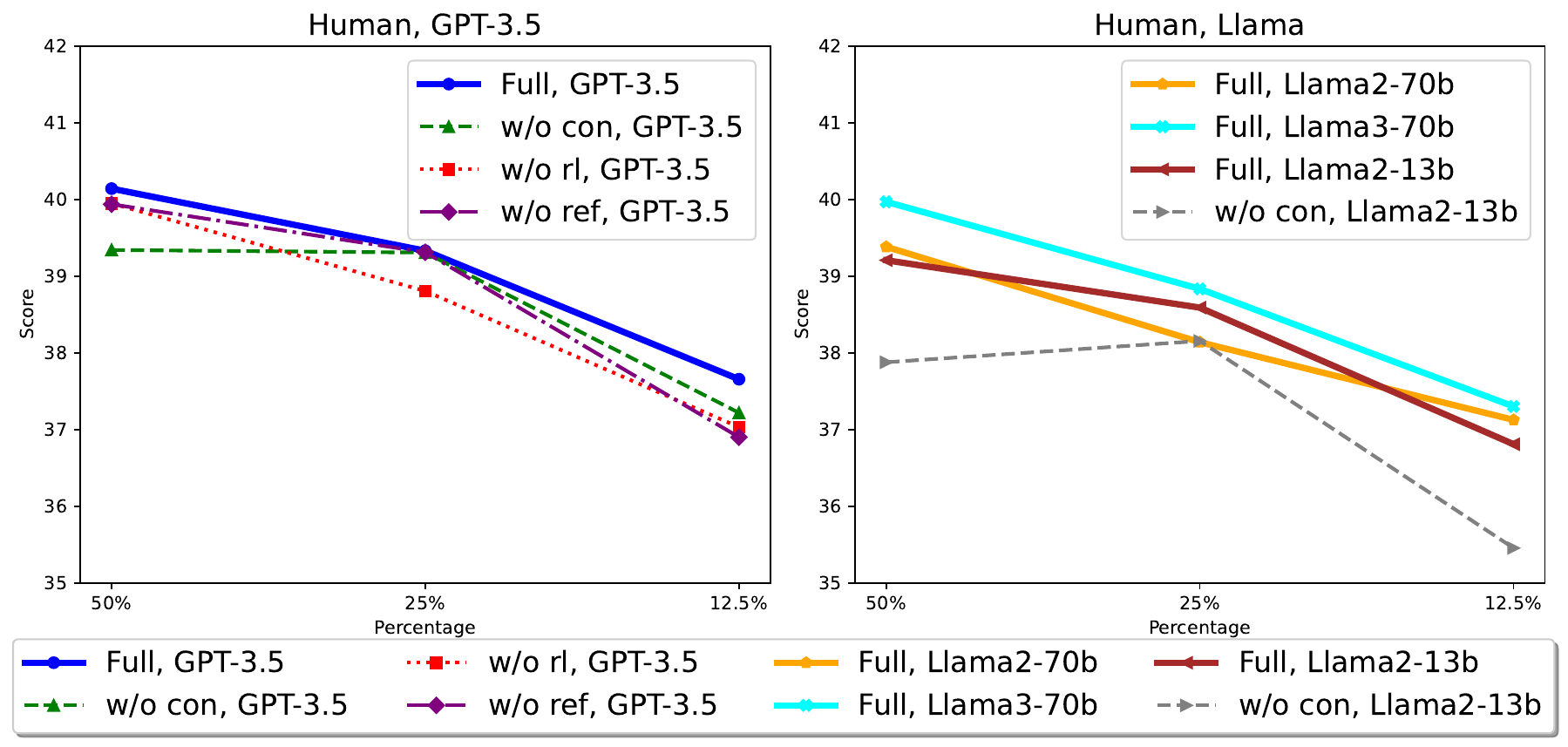}
\caption{Performance of LLMs fine-tuned on subsets of human-written data selected by SCAR(ID), trained with different configurations and synthetic data sources (e.g., GPT-3.5, Llama).}
    \vspace{-3mm}
    \label{fig:ablation_summary}
\end{figure}

\paragraph{Style Representation Learning.}
Figure~\ref{fig:ablation_summary} shows that removing the representation learning loss (w/o rl, GPT-3.5) or excluding ``referenced'' responses (w/o ref, GPT-3.5) only slightly reduces LLM performance in the code domain. The objective in Eq.~(\ref{eq:disentanglement}) is likely satisfied even without the loss because ``referenced'' responses provide an intermediate style during training, which is why we set a low coefficient (0.1) for this loss. However, excluding ``referenced'' responses significantly degrades performance in the open domain (Table~\ref{tab:open_llm_no_ref}, Appendix~\ref{app:no_ref_impact}) and disrupts the optimization of Eq.~(\ref{eq:disentanglement}). Table~\ref{tab:distance_analysis_embeddings}, Appendix~\ref{app:rep_analysis} further analyses the representation learning results. 

\paragraph{Data Quality Constraint.} Figure~\ref{fig:ablation_summary} (2nd) shows that removing the data quality constraint in Eq.~(\ref{eq:constraint}) significantly worsens the performance of LLMs fine-tuned on human-crowdsourced data when \SCAR{} is trained on lower-quality datasets, such as \llamaTwoThirteenB-generated responses (w/o con, Llama2-13b), compared to using the constraint (Full, Llama2-13b). In this case, \SCAR{} tends to select style-consistent but erroneous or unhelpful examples from LLM SFT data with varying quality (e.g. crowdsourced data). However, in other cases, removing the quality constraint has minimal impact on data selection performance.

\paragraph{LLMs for Generating \SCAR{} Training Data.}
Figure~\ref{fig:ablation_summary} shows that the choice between \textsc{Llama} and \gptThreeFive for generating synthetic training data for \SCAR{} has minimal impact when the trained ranker selects from human-written data. However, when selecting from mixed synthetic \gptThreeFive data for LLM SFT, using \textsc{Llama}-generated data to train the ranker leads to slightly lower LLM performance (Table~\ref{tab:coding_domain_ablation_expanded_synthetic} in Appendix~\ref{sec:ablation_results}). This disparity likely results from stylistic differences between \textsc{Llama} and \gptThreeFive-generated responses.

\section{Related Work}

\paragraph{Instruction-Tuning Data Selection.} Instruction-tuning trains LLMs to follow complex instructions in various contexts~\citep{wei2021finetuned,sanh2021multitask}. Data are sourced from human-curated examples~\citep{wang2022super,zhou2023lima} and LLM outputs~\citep{xu2023wizardlm,wang2022self}. Studies~\citep{zhou2023lima, chen2023alpagasus, li2024superfiltering, li2023quantity, lu2023instag,liumakes} show that smaller, high-quality datasets can outperform significantly larger ones in boosting LLM performance. LIMA uses expert human curation for stylistic consistency~\citep{zhou2023lima}, while AlpaGasus~\citep{chen2023alpagasus} utilizes LLMs to assess data quality. Other methods select effective examples based on Instruction Following Difficulty scores~\citep{li2024superfiltering, li2023quantity}, diversity metrics~\citep{lu2023instag,bukharin2023data}, or response length~\citep{zhaolong}.  

\paragraph{Automatic Authorship Detection.} Our method relates to authorship detection studies. Traditional authorship detection used lexical features like TTR, MTLD, and Flesch readability scores~\citep{tripto-etal-2023-hansen,zheng2023review}. Recent focus has shifted to distinguishing human and machine-generated texts using advanced neural networks to analyze styles at the corpus~\citep{mitchell2023detectgpt,su2023detectllm} or the sentence levels~\citep{zeng2024towards,zeng2023towards,wang2023seqxgpt,zengdetecting}. The studies~\citep{xu2024detecting,su2023detectllm,wang2023seqxgpt,mitchell2023detectgpt,wu2023llmdet}, like ours, show perplexity effectively differentiates between human and machine styles. 

\section{Conclusion}
Our empirical study demonstrates that among training datasets with responses of comparable helpfulness and correctness, those exhibiting higher consistency in two key response style elements, linguistic form and instructional surprisal, significantly enhance the performance of fine-tuned LLMs. Building on this insight, we propose \SCAR, a ranking method designed to identify and select stylistically consistent training data for LLM fine-tuning. Our experiments reveal that LLMs fine-tuned on carefully selected small subsets can outperform models trained on complete datasets, with \SCAR achieving superior performance using only 0.7\% of the original data in the best case. Furthermore, \SCAR consistently outperforms other data selection baselines across various LLM fine-tuning scenarios.
\section*{Limitations}
\subsection*{Discussion of Bias}
Reducing the training dataset size can potentially introduce biases. To address this concern, we discuss two types of bias: fairness bias and lexical diversity bias.

\paragraph{Fairness Bias.} Our experiments (Tables~\ref{tab:corrected_fairness_metrics_arrows} and~\ref{tab:regard_diff_mixed_and_human} at Appendix~\ref{app:bias_fairness}) show that LLMs fine-tuned with \SCAR-selected subsets produce responses with minimal levels of toxicity and sentiment polarity toward certain demographic and occupational groups. Overall, the fairness performance of models trained on \SCAR-selected data is \textbf{comparable to, or even better than,} models trained on full datasets or data selected using other methods. While fairness biases may persist, we argue that this issue is not unique to \SCAR but remains a broader challenge for all LLMs. Refining selection criteria to further mitigate these biases is a promising direction for future work. See Appendix~\ref{app:bias_fairness} for a detailed analysis.

\paragraph{Lexical Diversity Bias.} Table~\ref{tab:lexical_diversity} in the Appendix evaluates lexical diversity in instructions and responses separately using TTR and MTLD. Results show that \SCAR-selected instructions exhibit a slightly lower TTR compared to the full dataset and other subsets, with a more noticeable TTR reduction observed in responses. However, MTLD scores, which measure length-independent lexical richness, remain comparable across \SCAR-selected data, the full dataset, and subsets chosen by other baselines. This suggests that while \SCAR reduces surface-level lexical variation (reflected by TTR), it does not significantly affect the overall depth and richness of vocabulary (captured by MTLD) in instructions or responses.

Importantly, instruction-level diversity is more critical for LLM fine-tuning performance~\citep{lu2023instag, bukharin2023data}, and \SCAR-selected subsets retain this essential instruction diversity. The slight reduction in TTR does not pose a significant concern, as evidenced by \SCAR{}'s strong performance across our experiments. See Appendix~\ref{app:bias_lexical_diversity} for a detailed analysis.

Additionally, we note that \SCAR{} does not perform instruction deduplication. When datasets contain duplicate instructions with high-scoring responses, \SCAR{} may select multiple instances of these duplicates, resulting in the same high-quality instruction-response pairs appearing repeatedly in the selected subset. This repetition reduces the diversity of selected subsets. Therefore, \textbf{we highly recommend preprocessing datasets to remove duplicates before applying our method for optimal results}.

\subsection*{Discussion of Initial Data Pool}
Another potential limitation is the initial data pool to be selected. \SCAR{} is specifically designed to improve LLM SFT performance by selecting subsets from datasets with \textbf{style-inconsistent} responses. When responses are already style-consistent, such as all responses generated by a single LLM, \SCAR{}'s advantages become limited due to reduced stylistic variation for the ranker to distinguish between responses. Our experiments on coding tasks with all \gptThreeFive-generated responses (Table~\ref{tab:single_source_baselines} in Appendix~\ref{app:single_source_analysis}) demonstrate that SCAR achieves only marginal improvements over baseline methods in such scenarios, with performance differences typically within 1-3 percentage points.

We argue that style-inconsistent data are prevalent in real-world scenarios, including crowd-sourced human data and synthetic data collected from different sources, highlighting the practical benefits of \SCAR{}. Moreover, the insight that style elements of linguistic form and instructional surprisal within responses significantly impact LLM SFT performance may be even more important than the data selection method itself, as this understanding can inform future dataset curation practices.

\bibliography{anthology,custom}



\newpage
\appendix
\onecolumn


\begin{center}
    \textbf{Appendix Table of Contents}
\end{center}
{
    \setcounter{tocdepth}{2} 
    \startcontents 
    \printcontents{}{1}{}
}

\section{Implementation Details}
\label{app:implement_details}
\subsection{Model Training Configurations}
We fine-tune the \llamaThreeEightB and \codeLlamaSevenB models using LoRA, a parameter-efficient tuning method, on NVIDIA A100 GPUs to minimize computational costs. Both models undergo three training epochs with a learning rate of $2 \times 10^{-5}$, using a cosine learning rate scheduler and a warm-up ratio of 0.03. Training is performed with BF16 and TF32 precision modes enabled. For \llamaThreeEightB, we employ a single GPU with a batch size of 2, while for \codeLlamaSevenB, two GPUs are used with the same batch size, incorporating LoRA parameters set to $r = 8$ and $\alpha = 16$. For the OpenAI models, we adopt \textsc{gpt-3.5-turbo-0125} and \textsc{gpt-4-1106-preview} as our default configurations. We set the maximum input length for the LLMs to 2048 tokens.

The \SCAR{} ranker is trained with a learning rate of $2 \times 10^{-5}$ for up to 20 epochs, using early stopping based on validation performance. For code domain tasks, we utilize \textsc{CodeT5p-110M-Embedding}~\citep{wang2023codet5+} for contextual representation encoding, while for open-domain tasks, we employ \textsc{RoBERTa-Base}~\citep{liu2019roberta}. When curating \StackExchange{} examples for the ranker and LLM training, we ensure quality by selecting instructions containing code blocks and answers with ratings above 2.

\subsection{Prompt for Generating Referenced Response}
\label{app:prompt_reference}
The prompt used to rewrite the human response to generate the ``referenced'' response is as follows:

\begin{lstlisting}
### Reference Answer:
{human response}

### Background 
You are a knowledgeable AI assistant.
Above is the reference answer. Below is an instruction that describes a task. Given the reference answer, write a response that appropriately completes the request.
Please keep the semantics of the reference answer unchanged in your response, while pretending as if you have never seen the reference answer, when crafting your final response.

### Instruction:
{instruction}

### Response:
\end{lstlisting}
\subsection{Prompt for Generating Direct Response}
\label{app:prompt_direct}
The prompt instruction to generate ``direct'' response is as follows:

\begin{lstlisting}
### Background 
You are a knowledgeable AI assistant.
Below is an instruction that describes a task. Please write a response that appropriately completes the request.

### Instruction:
{instruction}

### Response:
\end{lstlisting}

\section{Extended Analysis of Style Effects on LLM Fine-Tuning Performance}

\subsection{Extended Analysis of LLM Performance on Coding Tasks}
Table~\ref{tab:coding_domain_expanded} presents the detailed results for the coding tasks mentioned in Table~\ref{tab:one_column_performance_comparison}, providing a comprehensive breakdown of the Pass@1 and Pass@10 metrics for each task, rather than just the average scores.

Table~\ref{tab:coding_domain_expanded} reveals that ``direct'' responses outperform ``referenced'' responses across most programming benchmarks, suggesting that generating answers without mirroring human semantic content yields better results for coding tasks. For instance, \gptThreeFive-generated ``direct'' achieves a Pass@1 of 33.00\% on the \HumanEval{}  benchmark, compared to 28.58\% for \gptThreeFive-generated ``referenced,'' and similar trends are observed across Java, JavaScript, and C++ benchmarks. Human responses also lag behind ``direct'' and ``referenced'' responses, indicating that synthetic data can offer better stylistic consistency, which can boost LLM SFT performance. \llamaTwoSeventyB performs notably better than its smaller counterpart, \llamaTwoThirteenB, showing a clear advantage due to larger model scale, though it still falls short of \gptThreeFive in most metrics, highlighting \gptThreeFive’s stronger coding capabilities. Interestingly, fine-tuned base LLMs perform particularly well in JavaScript, likely due to its simpler syntax and predictable patterns, which chat-LLMs like \gptThreeFive can easily understand and replicate, leading to high-quality training data. These findings highlight the effectiveness of ``direct'' responses and underscore the importance of data quality and style consistency in fine-tuning LLMs for code generation.

\begin{table}[ht!]
\centering
\resizebox{0.98\textwidth}{!}{
\begin{tabular}{l|c|ccc}
\toprule
\multirow{2}{*}{\begin{tabular}[c]{@{}l@{}}\textbf{Data Curation} \\ \textbf{Methods}\end{tabular}} & \textbf{\HumanEval{}} & \multicolumn{3}{c}{\textbf{\MultiPLE{}}} \\
\cdashline{2-5}
& \textbf{Python} & \textbf{Java} & \textbf{JavaScript} & \textbf{C++} \\
& \textbf{Pass@1 / Pass@10} & \textbf{Pass@1 / Pass@10} & \textbf{Pass@1 / Pass@10} & \textbf{Pass@1 / Pass@10} \\
\hline
\hline
Human Response & 23.45 / 39.99 & 27.13 / 39.14 & 30.14 / 47.39 & 25.52 / 40.00 \\
\hline
\gptThreeFive & & \multicolumn{3}{c}{} \\
\ \ \ \ Referenced & 28.58 / 52.64 & 29.46 / 41.91 & 33.53 / 50.84 & 27.70 / 42.17 \\
\ \ \ \ Direct & 33.00 / 51.48 & 29.38 / 42.03 & 33.19 / 51.72 & 28.45 / 43.27 \\
\hline
\llamaTwoSeventyB & & \multicolumn{3}{c}{} \\
\ \ \ \ Referenced & 31.64 / 45.58 & 29.09 / 40.59 & 31.79 / 49.20 & 26.77 / 40.74 \\
\ \ \ \ Direct & 33.62 / 48.18 & 30.23 / 41.79 & 32.91 / 50.24 & 26.80 / 41.05 \\
\hline
\llamaTwoThirteenB & & \multicolumn{3}{c}{} \\
\ \ \ \ Referenced & 23.88 / 43.31 & 27.58 / 37.92 & 29.90 / 47.72 & 25.09 / 38.67 \\
\ \ \ \ Direct & 28.32 / 40.99 & 24.67 / 36.41 & 28.88 / 45.65 & 23.81 / 36.96 \\
\bottomrule
\end{tabular}
}
\caption{Detailed performance comparison of fine-tuned \codeLlamaSevenB evaluated on \HumanEval{} (Python) and \MultiPLE{} (Java, JavaScript, C++) coding benchmarks. The LLMs are fine-tuned on training sets curated with different response generation strategies and LLMs. The data examples are further filtered based on the perplexity similarity between ``referenced'' and human responses, excluding those with significant deviation. Pass@1 and Pass@10 scores for each programming language are reported.}
\label{tab:coding_domain_expanded}
\end{table}

\subsection{Extended Stylometric Analysis}
\label{appx:full-style-analysis}
\paragraph{Evaluation Settings.} To quantitatively evaluate stylistic consistency across datasets, we employ six stylometric metrics that capture distinct aspects of linguistic form, the structural elements that shape response presentation independent of semantics. Specifically, these metrics measure key linguistic form elements: transitional and functional word usage measured by TTR and MTLD of functional words, tone assessed by Flesch score, sentence structure quantified through Average Sentence Length, punctuation patterns captured by Punctuation Frequency, and layout features such as headers and bullet points measured by Layout Feature Frequency. Together with perplexity for assessing instructional surprisal, these metrics provide a comprehensive framework for analyzing response styles:

\begin{table}[ht!]
\centering
\resizebox{\textwidth}{!}{
\begin{tabular}{l|cc|cc|cc|cc|cc|cc||cc}
\toprule
\multirow{2}{*}{\begin{tabular}[c]{@{}l@{}}\textbf{Data Curation} \\ \textbf{Methods}\end{tabular}}  & \multicolumn{2}{c|}{\textbf{TTR}} & \multicolumn{2}{c|}{\textbf{MTLD}} & \multicolumn{2}{c|}{\textbf{Avg. Sent. Len.}} & \multicolumn{2}{c|}{\textbf{Punct. Freq.}}  & \multicolumn{2}{c|}{\textbf{Flesch Score}} & \multicolumn{2}{c||}{\textbf{Avg. Layout Freq.}} & \multicolumn{2}{c}{\textbf{PPL$(y|x)$}} \\
\cline{2-15} 
& \textbf{Mean} & \textbf{Std.} & \textbf{Mean} & \textbf{Std.} & \textbf{Mean} & \textbf{Std.} & \textbf{Mean} & \textbf{Std.} & \textbf{Mean} & \textbf{Std.} & \textbf{Mean} & \textbf{Std.} & \textbf{Mean} & \textbf{Std.} \\
\hline
\hline
\multicolumn{15}{c}{\textbf{StackExchange}} \\
\hline
Human Response & 62.06 & 24.23 & 11.58 & 7.71 & 124.37 & 100.22 & 42.80 & 31.96 & 38.33 & 43.97 & 0.42 & 1.36 & 1.85 & 0.33 \\
\hline
\gptThreeFive & \multicolumn{2}{c|}{}  & \multicolumn{2}{c|}{} & \multicolumn{2}{c|}{} & \multicolumn{2}{c|}{} & \multicolumn{2}{c|}{} & \multicolumn{2}{c||}{} & \multicolumn{2}{c}{} \\
\ \ \ \ Referenced & 31.65 & 8.16 & 13.61 & 2.51 & 46.49 & 20.90 & 44.88 & 25.38 & 57.32 & 16.16 & 0.10 & 0.28 & 1.84 & 0.33 \\
\ \ \ \ Direct & 34.15 & 8.14 & 13.34 & 2.57 & 46.31 & 23.59 & 38.80 & 20.48 & 54.66 & 16.92 & 0.26 & 0.41 & 1.78 & 0.30 \\
\hline
\llamaTwoSeventyB & \multicolumn{2}{c|}{}  & \multicolumn{2}{c|}{} & \multicolumn{2}{c|}{} & \multicolumn{2}{c|}{} & \multicolumn{2}{c|}{} & \multicolumn{2}{c||}{} & \multicolumn{2}{c}{} \\
\ \ \ \ Referenced & 44.01 & 11.90 & 14.28 & 3.66 & 70.34 & 51.50 & 42.30 & 36.70 & 54.12 & 21.73 & 0.18 & 0.52 & 1.81 & 0.36 \\
\ \ \ \ Direct & 45.67 & 13.52 & 14.20 & 4.23 & 83.18 & 84.01 & 35.82 & 26.28 & 51.78 & 24.34 & 0.28 & 0.72 & 1.57 & 0.28 \\
\hline
\llamaTwoThirteenB & \multicolumn{2}{c|}{}  & \multicolumn{2}{c|}{} & \multicolumn{2}{c|}{} & \multicolumn{2}{c|}{} & \multicolumn{2}{c|}{} & \multicolumn{2}{c||}{} & \multicolumn{2}{c}{} \\
\ \ \ \ Referenced & 31.97 & 7.46 & 15.64 & 3.06 & 43.03 & 25.11 & 50.31 & 28.81 & 62.73 & 17.23 & 0.13 & 0.42 & 1.76 & 0.27 \\
\ \ \ \ Direct & 33.35 & 8.86 & 14.90 & 3.12 & 43.49 & 27.49 & 39.60 & 22.64 & 61.44 & 16.92 & 0.22 & 0.38 & 1.76 & 0.28 \\
\hline
\multicolumn{15}{c}{\textbf{LIMA}} \\
\hline
Human Response & 31.77 & 20.49 & 15.21 & 4.38 & 32.41 & 49.18 & 64.54 & 63.70 & 63.71 & 27.98 & 0.43 & 1.37 & 4.42 & 1.53 \\
\hline
\gptThreeFive & \multicolumn{2}{c|}{}  & \multicolumn{2}{c|}{} & \multicolumn{2}{c|}{} & \multicolumn{2}{c|}{} & \multicolumn{2}{c|}{} & \multicolumn{2}{c||}{} & \multicolumn{2}{c}{} \\
\ \ \ \ Referenced & 48.40 & 18.43 & 15.28 & 6.04 & 26.51 & 21.36 & 14.27 & 10.73 & 59.45 & 19.25 & 0.15 & 0.64 & 4.02 & 1.52 \\
\ \ \ \ Direct & 47.53 & 16.06 & 15.08 & 5.31 & 24.87 & 17.04 & 14.08 & 9.33 & 55.59 & 21.00 & 0.26 & 0.58 & 2.51 & 0.64 \\
\hline
\llamaTwoSeventyB & \multicolumn{2}{c|}{}  & \multicolumn{2}{c|}{} & \multicolumn{2}{c|}{} & \multicolumn{2}{c|}{} & \multicolumn{2}{c|}{} & \multicolumn{2}{c||}{} & \multicolumn{2}{c}{} \\
\ \ \ \ Referenced & 39.32 & 16.51 & 15.15 & 4.88 & 25.67 & 21.47 & 27.76 & 19.84 & 61.77 & 18.43 & 0.33 & 0.46 & 3.51 & 1.45 \\
\ \ \ \ Direct & 37.02 & 15.63 & 14.62 & 4.84 & 24.76 & 18.59 & 27.94 & 17.11 & 59.66 & 18.16 & 0.43 & 0.50 & 2.09 & 0.42 \\
\hline
\llamaTwoThirteenB & \multicolumn{2}{c|}{}  & \multicolumn{2}{c|}{} & \multicolumn{2}{c|}{} & \multicolumn{2}{c|}{} & \multicolumn{2}{c|}{} & \multicolumn{2}{c||}{} & \multicolumn{2}{c}{} \\
\ \ \ \ Referenced & 35.74 & 13.64 & 15.98 & 4.42 & 24.65 & 14.75 & 27.44 & 17.70 & 64.46 & 17.45 & 0.16 & 0.42 & 3.10 & 1.19 \\
\ \ \ \ Direct & 31.90 & 14.22 & 15.08 & 3.78 & 22.60 & 12.61 & 35.22 & 18.74 & 62.30 & 15.40 & 0.37 & 0.39 & 2.06 & 0.38 \\
\bottomrule
\end{tabular}
}
\caption{Comprehensive performance comparison of stylometric analysis across datasets using instructions from \StackExchange{} and \LIMA{}, paired with responses generated by human writers and various LLMs, presenting the average (Mean) and standard deviation (Std.) for six authorship detection metrics and Perplexity$(y|x)$.}
\label{tab:style_analysis_expand}
\end{table}

\paragraph{Linguistic Form Metrics:} 
\begin{enumerate}
    \item \textbf{Type-Token Ratio (TTR)~\cite{templin1957certain}:} Measures lexical diversity by calculating the ratio of unique words (types) to the total number of words (tokens) in a text. A higher TTR indicates greater lexical diversity.
    \item \textbf{Measure of Textual Lexical Diversity (MTLD)~\cite{mccarthy2005assessment}:} MTLD is less sensitive to text length compared to TTR. It computes the average length of sequential word strings that maintain a given TTR value, where higher MTLD scores suggest greater lexical diversity.
    \item \textbf{Average Sentence Length (Avg. Sent. Len.):} Calculates the average number of words per sentence, providing insights into the syntactic complexity of the text.
    \item \textbf{Punctuation Frequency (Punct. Freq.):} Computes the frequency of punctuation marks within each response, reflecting the density of punctuation usage.
    \item \textbf{Flesch Reading Ease Score (Flesch Score):} Assesses readability based on the average sentence length and the average number of syllables per word. Higher scores indicate greater readability.
    \item \textbf{Layout Feature Frequency (Avg. Layout Freq.):} Calculates the frequency of structural elements (bullet points, headers, bold text) per sentence, representing the consistency of formatting and organizational patterns.
\end{enumerate}

\paragraph{Instructional Surprisal Metric:} 
\begin{itemize}
    \item \textbf{Perplexity of $P(y|x)$:} Captures the overall response surprisal given the instruction. 
\end{itemize}


\paragraph{Discussion.} Table~\ref{tab:style_analysis_expand} presents the average and standard deviation (\textit{Std.}) of these metrics across responses from human-written and LLM-generated texts for both \LIMA{} and \StackExchange{} instructions. Our analysis reveals that LLM-generated responses consistently demonstrate higher stylistic consistency compared to human-written ones, with responses synthesized by \gptThreeFive and \textsc{Llama2} showing lower standard deviations across most metrics. This indicates greater consistency in functional word diversity, sentence length, punctuation usage, readability, and layout features. Furthermore, ``direct'' responses achieve higher consistency in response surprisal than ``referenced'' and human responses, as evidenced by their lower standard deviation values of perplexities. 

Notably, even the \LIMA{} dataset, despite being optimized and curated by human experts for style consistency, exhibits lower stylistic consistency in our metrics compared to LLM-synthesized datasets. These results highlight both the inherent challenge of achieving style consistency through manual curation and the significant potential of using LLMs to generate stylistically consistent training data.

In conclusion, our stylometric analysis quantitatively validates that LLM-synthesized datasets demonstrate superior stylistic consistency compared to human-written responses across most measured dimensions.




\subsection{Impact of Maintaining Instructional Surprisal Consistency in Referenced Responses on Stylometric Analysis and Model Performance}
\label{sec:appendix_perplexity_filtering}

\begin{table*}[ht!]
\centering
\resizebox{0.95\textwidth}{!}{
\begin{tabular}{l|ccc|ccc}
\toprule
 & \multicolumn{3}{c|}{\textbf{StackExchange (10k)}} & \multicolumn{3}{c}{\textbf{LIMA (1k)}}  \\
\cline{2-7}
  \multirow{3}{*}{\begin{tabular}[c]{@{}l@{}}\textbf{Data Curation} \\ \textbf{Methods}\end{tabular}} & 
  \multicolumn{1}{c}{\begin{tabular}[c]{@{}c@{}}\textbf{Stylometric} \\ \textbf{Analysis}\end{tabular}} &
  \multicolumn{1}{c}{\begin{tabular}[c]{@{}c@{}}\textbf{Data} \\ \textbf{Quality}\end{tabular}} & 
  \multicolumn{1}{c|}{\begin{tabular}[c]{@{}c@{}}\textbf{\codeLlamaSevenB} \\ \textbf{Performance}\end{tabular}} & 
  \multicolumn{1}{c}{\begin{tabular}[c]{@{}c@{}}\textbf{Stylometric} \\ \textbf{Analysis}\end{tabular}} & 
  \multicolumn{1}{c}{\begin{tabular}[c]{@{}c@{}}\textbf{Data} \\ \textbf{Quality}\end{tabular}} & 
  \multicolumn{1}{c}{\begin{tabular}[c]{@{}c@{}}\textbf{\llamaThreeEightB} \\ \textbf{Performance}\end{tabular}} \\
\cdashline{2-7}
 & \multicolumn{1}{c}{\begin{tabular}[c]{@{}c@{}}\textbf{Std. TTR} /\\ \textbf{Std. PPL}\end{tabular}}  & 
 \multicolumn{1}{c}{\begin{tabular}[c]{@{}c@{}}\textbf{Helpfulness} /\\ \textbf{Correctness}\end{tabular}} & 
 \multicolumn{1}{c|}{\begin{tabular}[c]{@{}c@{}}\textbf{Avg. Pass@1} /\\ \textbf{Avg. Pass@10}\end{tabular}} & 
 \multicolumn{1}{c}{\begin{tabular}[c]{@{}c@{}}\textbf{Std. TTR} /\\ \textbf{Std. PPL}\end{tabular}} & 
 \multicolumn{1}{c}{\begin{tabular}[c]{@{}c@{}}\textbf{Helpfulness} /\\ \textbf{Correctness}\end{tabular}} & 
 \textbf{L.C. WinRate} \\
\hline
\hline
Human Response & 22.27 / 1.41 & 3.34 / 3.57 & 31.65 / 46.63 & 19.54 / 8.01 & 4.32 / 4.37 & 2.29 \\
\hline
\multicolumn{1}{l|}{\gptThreeFive} & \multicolumn{3}{c|}{} & \multicolumn{3}{c}{}  \\ 

\ \ \ \ Referenced & 7.95 / 0.31 & 3.65 / 3.60 & 31.66 / 48.82 & 17.43 / 5.86 & 4.05 / 4.32 & 4.07 \\

\ \ \ \ Direct & 7.75 / 0.28 &  3.55 / 3.50 & 35.11 / 49.68 & 16.43 / 3.61 & 4.18 / 4.49 & 7.15 \\
\hline
\multicolumn{1}{l|}{\llamaTwoSeventyB} & \multicolumn{3}{c|}{} & \multicolumn{3}{c}{}\\ 

\ \ \ \ Referenced & 11.09 / 0.48  & 3.47 / 3.33  & 30.16 / 46.44 & 16.08 / 5.04 & 4.25 / 4.36 & 4.27 \\

\ \ \ \ Direct & 12.49 / 0.25  & 3.03 / 3.03  & 33.11 / 47.35 & 15.60 / 3.11 & 4.33 / 4.44 & 8.14 \\
\hline
\multicolumn{1}{l|}{\llamaTwoThirteenB} & \multicolumn{3}{c|}{} & \multicolumn{3}{c}{}\\ 

\ \ \ \ Referenced & 7.29 / 0.24  & 2.82 / 2.54  & 26.88 / 42.87 & 12.96 / 3.49 & 4.03 / 4.00 & 3.94 \\

\ \ \ \ Direct & 8.27 / 0.22  & 2.09 / 1.93  & 25.13 / 37.73 & 13.18 / 1.13 & 3.66 / 3.78 & 6.80 \\
\bottomrule
\end{tabular}
}
\caption{Performance comparison of \codeLlamaSevenB and \llamaThreeEightB fine-tuned on training sets curated using different methods and various LLMs, without applying surprisal-based instruction filtering, along with data quality and stylometric analysis metrics for the training sets.}
\label{tab:one_column_performance_comparison_original}
\end{table*}

In Section~\ref{sec:RQ2}, we used perplexity-based filtering to exclude instructions where the surprisal of ``Referenced'' responses significantly differed from that of human responses. Specifically, we excluded instructions where the PPL$(y|x)$ of at least one ``Referenced'' response exceeded thresholds of 0.15 or 2.5. This filtering process reduced the dataset to 944 instructions from \StackExchange{} and 407 instructions from \LIMA{}.

Table~\ref{tab:one_column_performance_comparison_original} highlights the impact of dataset size on LLM fine-tuning performance in the coding domain. For human responses, the average Pass@1 score across all four programming languages increased from 26.56 to 31.65 after adding more data. Notably, the official base model \codeLlamaSevenB achieves a Pass@1 score of 29.98, while \textsc{CodeLlama-7b-Instruct} achieves 34.8 on \HumanEval{} on \texttt{BigCodeLeaderboard}\footnote{\url{https://huggingface.co/spaces/bigcode/bigcode-models-leaderboard}}. In contrast, Table~\ref{tab:coding_domain_expanded} reports a significantly lower Pass@1 of 23.45, mainly due to the reduced dataset size (944 examples). \textbf{With sufficient data and effective selection strategies, the Pass@1 score on \HumanEval{}  for base \codeLlamaSevenB trained on human responses can reach 33, while synthetic responses can further boost performance to around 40, as shown in Tables~\ref{tab:coding_domain_baseline_human_expanded} and~\ref{tab:coding_domain_baseline_gpt_expanded}.} As achieving high model performance is not the primary goal in Section~\ref{sec:RQ2}, controlled filtering is essential for accurately analyzing variations in the instructional surprisal of responses and their impact on LLM fine-tuning.

A key observation from the stylometric analysis is the measurement of instructional surprisal through perplexity. Interestingly, Table~\ref{tab:one_column_performance_comparison_original} shows, without filtering, ``referenced'' responses exhibit greater surprisal consistency compared to human-written responses, particularly within the \StackExchange{} code data. This finding is somewhat counterintuitive, as one might expect ``referenced'' responses--rewritten versions of human responses--to closely mirror the surprisal consistency of their human counterparts. We hypothesize that this discrepancy arises because LLMs, even when explicitly instructed to semantically align closely with human responses, may introduce subtle variations that affect surprisal metrics.

While perplexity-based filtering is critical for achieving a more accurate analysis of LLM performance under varying stylistic consistency conditions, it was not used for our \SCAR{} training for the following reasons: i) Our goal is to learn a function that ranks responses based on style consistency. As shown in Table~\ref{tab:one_column_performance_comparison_original}, ``Direct'' responses already demonstrate higher stylistic consistency compared to ``Referenced'' and human responses, fulfilling the ranking objective without the need for additional filtering. ii) Filtering removes a substantial number of examples, which could negatively impact training performance by reducing the dataset size.

\subsection{Independence Tests of Linguistic Form and Instructional Surprisal}
\label{app:cmi_calculation}
In this section, we examine whether the linguistic form features of responses are correlated with instructional surprisal and whether linguistic form depends on instructions. Understanding these relationships is essential for justifying the design of our ranking model, which employs distinct structures to represent these two sets of features.
\paragraph{Independence Between Linguistic Form and Instructional Surprial.}





To validate the independence between linguistic form and instructional surprisal, we conduct two complementary analyses:

\textbf{\textit{Regression Analysis:}}  
We perform regression modeling on the \LIMA{} dataset to predict the instructional surprisal metric, perplexity $PPL(y|x)$, based on two feature sets:
\begin{itemize}
    \item \textbf{Linguistic form features:} unigrams of functional words, TTR and MTLD of functional words, punctuation and layout patterns, and Flesch readability scores. 
    \item \textbf{Semantic features:} contextual token embeddings derived from \textsc{sentence-transformers/all-MiniLM-L6-v2}~\cite{reimers2019sentence}, a model pre-trained for semantic encoding and paraphrase detection tasks. 
\end{itemize}

The average absolute regression coefficients indicate that semantic features are significantly more influential in predicting instructional surprisal, with an average importance score of 1.193, compared to only 0.236 for each linguistic form feature.


\textbf{\textit{Variance Analysis.}}
We further investigate the independence of linguistic form and instructional surprisal by analyzing variance patterns in $PPL(y|x)$. Responses are decomposed into semantic tokens ($y_c$) and functional non-semantic tokens ($y_p$), which represent a key component of linguistic form elements (see Section~\ref{app:semantic_nonsemantic_extraction} for token separation details). By comparing the variance contributions of $PPL(y_c|y_p, x)$ and $PPL(y_p|y_c, x)$ to $PPL(y|x)$, we find: 
\begin{itemize}
    \item \textbf{Semantic tokens ($y_c$):} explain 283.67\% of the variance.  
    \item \textbf{Functional tokens ($y_p$):} explain only 4.01\% of the variance.  
\end{itemize}

The combined evidence from our regression and variance analyses suggests that linguistic form and instructional surprisal are distinct dimensions of response style, with only a weak correlation between them. Semantic features are the primary contributors to instructional surprisal, with linguistic form playing a much smaller role.

\paragraph{Independence Tests between Linguistic Form and Instructions}

We employ Conditional Mutual Information (CMI)~\citep{WYNER197851} to quantify the dependencies between semantic tokens ($y_c$) and non-semantic tokens ($y_p$) with respect to instructions ($x$). For semantic content and instructions, CMI is defined as:
\[
I(y_c; x \mid y_p) = \frac{1}{N} \sum_{i=1}^{N} \log \left( \frac{P(y_c^{(i)} \mid x^{(i)}, y_p^{(i)})}{P(y_c^{(i)} \mid y_p^{(i)})} \right),
\]
with an analogous formulation for functional tokens:
\[
I(y_p; x \mid y_c) = \frac{1}{N} \sum_{i=1}^{N} \log \left( \frac{P(y_p^{(i)} \mid x^{(i)}, y_c^{(i)})}{P(y_p^{(i)} \mid y_c^{(i)})} \right).
\]
Using \llamaThreeEightB to estimate conditional probabilities and a POS-based approach to separate semantic and non-semantic functional tokens (detailed in Appendix~\ref{app:semantic_nonsemantic_extraction}), we analyze both human-written and \gptThreeFive-generated responses with \LIMA{} and \StackExchange{} instructions.

For \LIMA{} instructions, the mutual information scores reveal that semantic tokens show a stronger dependence on instructions, with $I(y_c; x \mid y_p) = 0.4$, compared to $I(y_p; x \mid y_c) = 0.15$. Similarly, for \StackExchange{} instructions, semantic tokens again dominate with $I(y_c; x \mid y_p) = 0.49$, while functional tokens exhibit a much weaker dependence at $I(y_p; x \mid y_c) = 0.03$. Since functional tokens are key indicators of linguistic form, these findings confirm that linguistic form has a significantly weaker dependence on instructions compared to semantic tokens. Therefore, in Eq.~(\ref{eq:ranker_model}), we aim to use max pooling over their representations to capture linguistic form features as non-semantic surface characteristics of responses without explicitly modeling their relationship to the instruction. This approach aligns with our findings, indicating that linguistic form is only weakly correlated with instructional context and has minimal impact on instructional surprisal.

\subsection{Background on Instructional Surprisal}
\label{app:surprisal_modelling}

Surprisal, traditionally defined as the negative log-probability of a word given its preceding context,
\[
-\log P(w \mid \text{context}),
\]
is a well-established indicator of cognitive processing difficulty and neural activation, including the N400 ERP component~\citep{10.1162/tacl_a_00548, goodkind2018predictive, michaelov2023strong, karampiperis2014towards}. This word-level metric quantifies how unexpected a word is given its context and is naturally derived from autoregressive language models trained on next-token prediction.

\paragraph{Extending Surprisal to Instruction-Level Evaluation.}
At the sequence level, surprisal can be generalized to assess the probability of an entire response $W$ given an instruction:
\begin{equation}
P(W \mid \text{instruction}) = \prod_{i=1}^N P(w_i \mid w_1, \ldots, w_{i-1}, \text{instruction}),
\label{eq:surprisal_sent}
\end{equation}
where $w_i$ denotes the $i$-th token in $W$, and $N$ is the length of the response. Based on this formulation, we define \textbf{instructional surprisal} as the surprisal of a full response conditioned on its instruction, reflecting the response's predictability under the language model.

\paragraph{Approaches to Modeling Instructional Surprisal.}
We consider two main approaches to estimating instructional surprisal: perplexity and semantic relatedness.

\textbf{Perplexity (PPL)} is a widely-used metric derived from the average surprisal of each token in a sequence. It is computed as:
\[
\text{PPL}(W) = \exp\left( -\frac{1}{N} \sum_{i=1}^N \log P(w_i \mid w_1, \ldots, w_{i-1}) \right),
\]
which corresponds to the exponentiated average negative log-likelihood. Perplexity thus provides a global measure of the predictability of a response. Since perplexity is a monotonic transformation of sequence-level surprisal (Eq.~\ref{eq:surprisal_sent}), it serves as a proxy for instructional surprisal.

\textbf{Semantic Relatedness} captures the semantic alignment between an instruction and its response. It reflects how topically and conceptually coherent the two are~\citep{salicchi2023study}. While originally proposed for word-level prediction, prior work has used semantic similarity between a word vector $\vec{w}$ and its context vector $\vec{h}$, often computed via cosine similarity, to estimate semantic surprisal~\citep{sayeed2015vector}:
\begin{equation}
\text{Surprisal}(w \mid h) = -\log P(w \mid h), \quad \text{where} \quad P(w \mid h) \propto \cos(\vec{w}, \vec{h}) = \frac{\vec{w} \cdot \vec{h}}{\|\vec{w}\| \|\vec{h}\|}.
\label{eq:relatedness_surprisal}
\end{equation}
In our case, we adapt this idea to the instruction-response level by using sentence embeddings in place of word vectors. However, such linear approaches may fail to fully capture complex, non-linear semantic dependencies between instruction and response.

Please note that while prior studies have reported significant correlations between semantic relatedness and surprisal~\citep{salicchi2023study,michaelov2023strong}, and some~\cite{sayeed2015vector} even estimate surprisal directly from semantic relatedness as in Eq.(\ref{eq:relatedness_surprisal}), other work has highlighted important distinctions between the two~\citep{salicchi2023study}.


\paragraph{Our Approach: Non-Linear Semantic Modeling via \SCAR.}
To overcome the limitations of conventional methods, \SCAR adopts a more expressive modeling strategy. Rather than relying on perplexity or simple embedding distances, it leverages a \textit{Relation Network}~\citep{sung2018learning} implemented as a multilayer perceptron ($\text{MLP}_c$, Eq.~\ref{eq:ranker_model}) to learn rich, non-linear alignment patterns between instructions and responses.

This design offers several advantages:

\begin{itemize}
    \item \textbf{Preservation of Surprisal Semantics:} \SCAR is trained using a triplet loss (Eq.~\ref{eq:disentanglement}) that enforces a structured representation space. Pairs with similar surprisal values are encouraged to lie closer in the embedding space, allowing the network to learn a disentangled representation that captures both linguistic form and instructional surprisal.
    
    \item \textbf{End-to-End Differentiability:} Unlike perplexity, which must be computed using non-differentiable external LLMs, the \textsc{RoBERTa}-based encoder and MLP components in \SCAR can be jointly optimized with downstream objectives, enabling unified training across the entire ranking model.
    
    \item \textbf{Practical Efficiency:} Token-level perplexity computation with large LLMs (e.g., \textsc{Llama-3-8B}) is computationally intensive and unsuitable for large-scale filtering. In contrast, \SCAR's relatedness scores are computed with a single forward pass through a lightweight transformer and MLP, offering a more scalable alternative.
\end{itemize}

\subsection{Identification of Semantic and Non-Semantic (Functional) Words}
\label{app:semantic_nonsemantic_extraction}

To distinguish between semantic content ($y_c$) and non-semantic (linguistic form-related) words ($y_p$) in the responses, we adopt a heuristic approach based on part-of-speech (POS) tagging. Specifically, content words--nouns, verbs, adjectives, and adverbs--are classified as semantic, while other POS tags (e.g., pronouns, conjunctions, prepositions, and determiners) are categorized as non-semantic. 

For code-related responses, we also treat code blocks as semantic content, given their integral role in conveying the main content of the response. Code blocks are identified using regular expressions that capture common code delimiters, such as triple backticks (\verb|```|), tildes (\verb|~~~|), and inline code marked by single backticks (\verb|`|).

Given the limitations of current NLP techniques, achieving perfect separation between semantic and non-semantic elements is challenging. However, our primary goal is not absolute precision, but to perform independence tests on various stylistic features relative to instructions and estimate instructional surprisal to inform our data selection ranker design. By focusing on comparative patterns, our approach effectively captures the impact of semantic and non-semantic content on stylistic consistency, and how these patterns influence data selection, ultimately improving LLM alignment through SFT.

To illustrate, Table~\ref{tab:semantic_nonsemantic_example} provides an example of how a response is split into semantic and non-semantic content using this method.

\begin{table}[ht!]
\centering
\begin{tabular}{|p{0.95\textwidth}|}
\hline
\textbf{Instruction} \\
\hline
Why is FIFA against adding instant replay to the game? \\
\hline
\textbf{Response} \\
\hline
{\color{blue}FIFA} {\color{black}is} {\color{blue}against} {\color{blue}adding} {\color{blue}instant} {\color{blue}replay} {\color{black}to} {\color{black}the} {\color{blue}game} {\color{black}because} {\color{black}they} {\color{blue}believe} {\color{black}it} {\color{black}would} {\color{blue}disrupt} {\color{black}the} {\color{blue}flow} {\color{black}of} {\color{black}the} {\color{blue}game} {\color{black}and} {\color{blue}potentially} {\color{blue}lead} {\color{black}to} {\color{blue}delays}. {\color{black}They} {\color{blue}also} {\color{blue}argue} {\color{black}that} {\color{blue}human} {\color{blue}error} {\color{black}is} {\color{black}a} {\color{blue}part} {\color{black}of} {\color{black}the} {\color{blue}game} {\color{black}and} {\color{blue}adding} {\color{blue}instant} {\color{blue}replay} {\color{black}could} {\color{blue}take} {\color{blue}away} {\color{black}from} {\color{black}the} {\color{blue}excitement} {\color{black}and} {\color{blue}unpredictability} {\color{black}of} {\color{blue}live} {\color{blue}sports}. Additionally, {\color{blue}implementing} {\color{blue}instant} {\color{blue}replay} {\color{black}would} {\color{blue}require} {\color{blue}significant} {\color{blue}changes} {\color{black}to} {\color{black}the} {\color{blue}rules} {\color{black}and} {\color{blue}regulations} {\color{black}of} {\color{black}the} {\color{blue}game}, {\color{black}which} {\color{blue}FIFA} {\color{black}may} {\color{black}be} {\color{blue}hesitant} {\color{black}to} {\color{blue}do}. \\
\hline
\end{tabular}
\caption{Visualization of semantic and non-semantic words selected based on the POS tags in the response. Semantic words are in {\color{blue}blue} and functional words are in black.}
\label{tab:semantic_nonsemantic_example}
\end{table}

\section{Extended Analysis of Main Experiments}

\subsection{Extended Analysis of Evaluation Results for Data Selection in Human-Written Coding Data}
\begin{table}[ht!]
\centering
\resizebox{0.98\textwidth}{!}{
\begin{tabular}{l|c|ccc}
\toprule
\multirow{2}{*}{\begin{tabular}[c]{@{}l@{}}\textbf{Data Sampling} \\ \textbf{Methods}\end{tabular}} & \textbf{\HumanEval} & \multicolumn{3}{c}{\textbf{\MultiPLE}} \\
\cdashline{2-5}
& \textbf{Python} & \textbf{Java} & \textbf{JavaScript} & \textbf{C++} \\
& \textbf{Pass@1 / Pass@10} & \textbf{Pass@1 / Pass@10} & \textbf{Pass@1 / Pass@10} & \textbf{Pass@1 / Pass@10} \\
\hline
\hline
Full Data & 32.87 / 48.24 & 30.92 / 44.92 & 33.84 / 52.62 & 28.51 / 43.91 \\
\hline
\SCAROOD & & \multicolumn{3}{c}{} \\
\ \ \ \ 50\%  & 31.94 / 47.80 & 30.85 / 43.29 & 33.91 / 52.45 & 29.23 / 45.28 \\
\ \ \ \ 25\%  & 31.85 / 46.80 & 29.97 / 43.24 & 33.14 / 52.75 & 29.20 / 45.21 \\
\ \ \ \ 12.5\%  & 30.77 / 46.80 & 28.92 / 41.86 & 31.23 / 48.38 & 28.17 / 43.61 \\
\hline
\SCARID & & \multicolumn{3}{c}{} \\
\ \ \ \ 50\%  & 33.83 / 50.24 & 30.10 / 44.95 & 34.46 / 53.10 & 28.25 / 43.71 \\
\ \ \ \ 25\%  & 31.48 / 48.68 & 30.76 / 44.60 & 32.91 / 52.15 & 28.92 / 43.98 \\
\ \ \ \ 12.5\%  & 31.10 / 47.14 & 29.46 / 43.06 & 31.38 / 49.11 & 27.61 / 42.39 \\

\hline
\Random & & \multicolumn{3}{c}{} \\
\ \ \ \ 50\%  & 29.79 / 44.06 & 30.14 / 43.90 & 32.86 / 51.61 & 28.48 / 43.89 \\
\ \ \ \ 25\%  & 30.04 / 45.76 & 30.22 / 42.35 & 33.06 / 51.05 & 28.89 / 43.89 \\
\ \ \ \ 12.5\%  & 27.94 / 45.79 & 27.53 / 40.47 & 31.48 / 51.25 & 25.29 / 40.51 \\

\hline
\Perplexity & & \multicolumn{3}{c}{} \\
\ \ \ \ 50\%  & 33.27 / 47.90 & 29.73 / 42.16 & 32.67 / 52.13 & 28.46 / 43.40 \\
\ \ \ \ 25\%  & 32.29 / 47.05 & 29.33 / 42.40 & 32.45 / 50.10 & 28.73 / 44.78 \\
\ \ \ \ 12.5\%  & 27.40 / 45.13 & 28.67 / 40.77 & 31.30 / 50.71 & 26.36 / 41.75 \\
\hline
\Superfiltering & & \multicolumn{3}{c}{} \\
\ \ \ \ 50\%  & 26.50 / 42.00 & 29.72 / 43.53 & 32.97 / 52.40 & 27.86 / 44.86 \\
\ \ \ \ 25\%  & 24.12 / 38.51 & 29.29 / 42.76 & 32.50 / 53.20 & 26.89 / 41.01 \\
\ \ \ \ 12.5\%  & 8.22 / 25.58 & 26.79 / 38.83 & 30.11 / 49.20 & 23.99 / 36.82 \\
\hline
\HFR & & \multicolumn{3}{c}{} \\
\ \ \ \ 50\%  & 20.29 / 41.52 & 30.41 / 44.11 & 33.49 / 51.27 & 28.71 / 44.83 \\
\ \ \ \ 25\%  & 11.20 / 25.73 & 29.38 / 42.81 & 31.73 / 51.51 & 28.09 / 43.07 \\
\ \ \ \ 12.5\%  & 11.04 / 27.74 & 27.51 / 40.82 & 30.71 / 49.41 & 24.91 / 39.77 \\
\hline
\AlpaGasus & & \multicolumn{3}{c}{} \\
\ \ \ \ 50\%  & 31.30 / 44.90 & 30.59 / 43.41 & 34.21 / 52.48 & 29.45 / 43.91 \\
\ \ \ \ 25\%  & 30.32 / 45.00 & 29.73 / 42.78 & 32.24 / 51.65 & 28.29 / 44.15 \\
\ \ \ \ 12.5\%  & 24.76 / 41.90 & 28.24 / 42.12 & 30.84 / 49.56 & 26.17 / 41.12 \\
\hline
\Diversity & & \multicolumn{3}{c}{} \\
\ \ \ \ 50\%  & 33.05 / 48.38 & 30.53 / 44.06 & 34.02 / 53.99 & 28.84 / 42.60 \\
\ \ \ \ 25\%  & 30.38 / 44.52 & 30.04 / 42.53 & 33.34 / 52.71 & 28.68 / 44.66 \\
\ \ \ \ 12.5\%  & 25.87 / 44.07 & 27.35 / 39.37 & 30.48 / 49.65 & 24.99 / 40.38 \\
\hline
\Longest & & \multicolumn{3}{c}{} \\
\ \ \ \ 50\%  & 30.99 / 50.90 & 30.74 / 44.74 & 32.17 / 52.47 & 28.32 / 43.55 \\
\ \ \ \ 25\%  & 30.10 / 48.41 & 29.35 / 42.65 & 30.72 / 51.98 & 28.92 / 45.07 \\
\ \ \ \ 12.5\%  & 28.12 / 47.60 & 28.54 / 41.97 & 29.53 / 48.43 & 27.40 / 41.65 \\
\bottomrule
\end{tabular}
}
\caption{Detailed performance comparison of fine-tuned \codeLlamaSevenB evaluated on the \HumanEval (Python) and \MultiPLE (Java, JavaScript, C++) coding benchmarks. The models are fine-tuned on human-written datasets selected with different selection methods and proportions. The table reports Pass@1 and Pass@10 scores for each individual programming language.}
\label{tab:coding_domain_baseline_human_expanded}
\end{table}

Table~\ref{tab:coding_domain_baseline_human_expanded} offers a comprehensive breakdown of LLM performance when fine-tuned on datasets sampled using various data selection strategies, expanding upon the average results presented in Figure~\ref{fig:main_res}. While the figure provides aggregated metrics, this table delivers a detailed view of Pass@1 and Pass@10 scores for each programming language across the \HumanEval{}  and \MultiPLE{}  benchmarks. This detailed presentation highlights performance variations in Python, Java, JavaScript, and C++.

The performance ranking of data selection methods aligns consistently with the trends shown in Figure~\ref{fig:main_res}, reinforcing our findings' reliability. Strategies such as \SCARID{} and Perplexity-based sampling demonstrate robust performance across most languages, while approaches like \HFR{} and \Superfiltering{} yield less favourable results, particularly with smaller data proportions. Notably, LLMs trained on our \SCARID{}-selected data outperform those trained on the full dataset when the selection portion exceeds 25\%, highlighting the superiority of our method. This result indicates that a carefully curated subset can sometimes produce better outcomes than using the entire dataset.

For a detailed explanation of the Pass@1 and Pass@10 metrics, please refer to the \HumanEval{}  paper by~\citet{chen2021codex}.

\subsection{Extended Analysis of Evaluation Results for Data Selection in Mixed Synthetic Coding Data}
\begin{table}[ht!]
\centering
\resizebox{0.98\textwidth}{!}{
\begin{tabular}{l|c|ccc}
\toprule
\multirow{2}{*}{\begin{tabular}[c]{@{}l@{}}\textbf{Data Sampling} \\ \textbf{Methods}\end{tabular}} & \textbf{\HumanEval} & \multicolumn{3}{c}{\textbf{\MultiPLE}} \\
\cdashline{2-5}
& \textbf{Python} & \textbf{Java} & \textbf{JavaScript} & \textbf{C++} \\
& \textbf{Pass@1 / Pass@10} & \textbf{Pass@1 / Pass@10} & \textbf{Pass@1 / Pass@10} & \textbf{Pass@1 / Pass@10} \\
\hline
\hline
Full Data & 40.63 / 54.93 & 32.67 / 44.24 & 36.89 / 54.10 & 32.68 / 45.65 \\
\hline
\SCAROOD & & \multicolumn{3}{c}{} \\
\ \ \ \ 50\%  & 40.15 / 55.25 & 32.15 / 44.44 & 37.01 / 55.59 & 31.96 / 46.59 \\
\ \ \ \ 25\%  & 38.23 / 52.58 & 32.57 / 45.44 & 37.04 / 53.20 & 30.60 / 45.67 \\
\ \ \ \ 12.5\%  & 38.29 / 52.74 & 32.46 / 45.45 & 36.07 / 53.45 & 31.91 / 45.56 \\
\hline
\SCARID & & \multicolumn{3}{c}{} \\
\ \ \ \ 50\%  & 40.98 / 56.57 & 32.80 / 45.75 & 37.58 / 55.69 & 32.73 / 45.71 \\
\ \ \ \ 25\%  & 39.84 / 56.75 & 32.52 / 43.83 & 36.67 / 55.32 & 32.00 / 46.26 \\
\ \ \ \ 12.5\%  & 36.93 / 52.96 & 32.62 / 44.82 & 36.45 / 52.33 & 30.43 / 45.42 \\

\hline
\Random & & \multicolumn{3}{c}{} \\
\ \ \ \ 50\%  & 39.04 / 51.80 & 31.75 / 44.85 & 35.59 / 55.13 & 32.76 / 46.34 \\
\ \ \ \ 25\%  & 35.61 / 52.40 & 31.33 / 44.24 & 36.68 / 54.23 & 30.53 / 44.60 \\
\ \ \ \ 12.5\%  & 34.99 / 51.90 & 31.34 / 44.29 & 35.91 / 51.63 & 31.08 / 44.49 \\

\hline
\Perplexity & & \multicolumn{3}{c}{} \\
\ \ \ \ 50\%  & 31.91 / 50.94 & 32.44 / 45.37 & 37.02 / 54.75 & 33.22 / 46.19 \\
\ \ \ \ 25\%  & 35.55 / 48.65 & 31.85 / 45.44 & 35.40 / 51.75 & 31.28 / 43.32 \\
\ \ \ \ 12.5\%  & 27.37 / 43.06 & 30.90 / 44.19 & 36.34 / 48.74 & 30.46 / 42.96 \\
\hline
\Superfiltering & & \multicolumn{3}{c}{} \\
\ \ \ \ 50\%  & 38.93 / 54.55 & 31.80 / 44.48 & 35.03 / 54.40 & 32.22 / 47.25 \\
\ \ \ \ 25\%  & 35.93 / 51.41 & 32.47 / 44.10 & 34.46 / 53.13 & 30.89 / 44.90 \\
\ \ \ \ 12.5\%  & 34.35 / 49.81 & 30.34 / 42.81 & 32.97 / 50.60 & 30.46 / 44.22 \\
\hline
\HFR & & \multicolumn{3}{c}{} \\
\ \ \ \ 50\%  & 39.09 / 53.59 & 32.42 / 43.90 & 36.11 / 53.51 & 31.60 / 45.51 \\
\ \ \ \ 25\%  & 38.04 / 53.36 & 32.57 / 43.51 & 36.45 / 54.10 & 31.27 / 46.28 \\
\ \ \ \ 12.5\%  & 29.20 / 50.06 & 31.87 / 43.85 & 35.17 / 53.94 & 30.02 / 44.31 \\
\hline
\AlpaGasus & & \multicolumn{3}{c}{} \\
\ \ \ \ 50\%  & 36.88 / 53.05 & 32.20 / 45.65 & 36.57 / 54.84 & 33.07 / 45.77 \\
\ \ \ \ 25\%  & 32.52 / 49.55 & 31.37 / 42.82 & 33.32 / 51.72 & 30.37 / 44.69 \\
\ \ \ \ 12.5\%  & 29.08 / 45.07 & 31.09 / 43.09 & 34.82 / 52.53 & 29.73 / 44.16 \\
\hline
\Diversity & & \multicolumn{3}{c}{} \\
\ \ \ \ 50\%  & 39.21 / 54.95 & 32.10 / 45.48 & 37.25 / 54.58 & 32.60 / 46.33 \\
\ \ \ \ 25\%  & 35.29 / 51.33 & 32.00 / 43.41 & 36.10 / 55.44 & 30.98 / 45.19 \\
\ \ \ \ 12.5\%  & 33.60 / 50.18 & 31.78 / 44.92 & 34.82 / 51.92 & 30.91 / 44.10 \\
\hline
\Longest & & \multicolumn{3}{c}{} \\
\ \ \ \ 50\%  & 36.83 / 53.90 & 32.73 / 45.15 & 36.73 / 55.92 & 33.85 / 46.83 \\
\ \ \ \ 25\%  & 35.60 / 53.50 & 32.34 / 45.54 & 36.25 / 54.65 & 32.57 / 46.43 \\
\ \ \ \ 12.5\%  & 34.54 / 49.89 & 32.41 / 46.31 & 35.57 / 54.64 & 31.42 / 45.30 \\

\bottomrule
\end{tabular}
}
\caption{Detailed performance comparison of fine-tuned \codeLlamaSevenB evaluated on the \HumanEval (Python) and \MultiPLE (Java, JavaScript, C++) coding benchmarks. The models are all fine-tuned using \gptThreeFive-generated datasets selected with different data selection methods and varying proportions. The table reports the Pass@1 and Pass@10 scores for each individual programming language.}
\label{tab:coding_domain_baseline_gpt_expanded}
\end{table}
Table~\ref{tab:coding_domain_baseline_gpt_expanded} offers a detailed breakdown of the LLM performance results summarized in Figure~\ref{fig:main_res}. It presents Pass@1 and Pass@10 scores across four programming languages, evaluating LLMs fine-tuned on synthetic dataset subsets chosen through various selection methods. This comprehensive view provides insights into the LLM's performance on individual tasks and programming languages, complementing the aggregated results shown in the figure.

\subsection{Extended Analysis of Evaluation Results for Open-Domain Data Selection Experiments}

\begin{table}[ht!]
\centering
\resizebox{\textwidth}{!}{
\begin{tabular}{l|ccccccccc}
\toprule
 & \multicolumn{9}{c}{\textbf{Methods}} \\

 &  \textbf{\SCARID} & \textbf{\SCAROOD} & \textbf{\Random} & \textbf{\Perplexity} & \textbf{\Superfiltering} & \textbf{\HFR} & \textbf{\AlpaGasus}& \textbf{\Diversity}& \textbf{\Longest}\\
\midrule
\textbf{Human} &  \multicolumn{9}{c}{} \\
\ \ \ \ 100\% &   \multicolumn{9}{c}{2.34} \\
\ \ \ \ 50\% &   2.24 & 1.90 & 2.03 & 1.74 & 2.00 & 1.50 & 2.09 & 1.99 & 1.46 \\
\ \ \ \ 25\% &   2.43 & 2.59 & 1.92 & 2.12 & 1.82 & 1.66 & 1.83 & 1.97 & 1.75\\
\ \ \ \ 10\% &   2.67 & 2.02 & 2.13 & 2.51 & 2.04 & 2.21 & 1.96 & 2.03 & 1.27 \\
\midrule
\textbf{Synthetic}  & \multicolumn{9}{c}{} \\
\ \ \ \ 100\% &   \multicolumn{9}{c}{3.64} \\
\ \ \ \ 50\% &  5.56 & 5.31 & 2.61 & 4.17 & 4.22 & 3.86 & 3.86 & 3.56 & 6.29 \\
\ \ \ \ 25\% &  5.89 & 5.08 & 3.00 & 4.04 & 5.70 & 4.30 & 3.94 & 2.51 & 5.32 \\
\ \ \ \ 10\% &  6.61 & 4.94 & 2.38 & 4.54 & 5.38 & 4.06 & 4.78 & 3.02 & 6.61 \\
\bottomrule
\end{tabular}
}
\caption{Detailed comparison of Length Control WinRate for fine-tuned \llamaThreeEightB models evaluated on \AlpacaEval benchmarks. Models are trained using human-written and synthetic \gptThreeFive-generated data, sampled with various selection methods and proportions.}
\label{tab:general_domain_expanded}
\end{table}

Table~\ref{tab:general_domain_expanded} presents the detailed numerical values for the Length Control WinRate, complementing the visual representation provided in Figure~\ref{fig:main_res}. The results show that for the selection of human data, \SCARID{} and \SCAROOD{} achieve competitive performance even at reduced data proportions, with \SCARID{} showing a slight advantage as the data size decreases, especially at the 25\% and 10\% subsets. In contrast, methods such as \Random and \HFR struggle to maintain consistently high performance across different data scales.

For the selection of synthetic \gptThreeFive-generated data, \SCARID{} consistently outperforms all methods except Longest, with WinRates peaking at 6.61 for the 10\% subset. Interestingly, Longest performs comparably to \SCARID{} when selecting synthetic data, as it tends to favour \textsc{Evol-Instruct}-generated data, which produces longer responses. This finding highlights that response token length can serve as a strong stylistic indicator, aligning with the principles of our style consistency framework.

These results suggest that well-curated synthetic datasets can enable high-performing chat-LLMs even at significantly reduced data proportions. Furthermore, traditional methods such as Random and Perplexity exhibit lower performance, underscoring the importance of selection strategies tailored to stylistic consistency in synthetic data scenarios. Striking a balance between data size, diversity, and style consistency remains crucial for optimizing performance.


\subsection{Extended Analysis of Style and Quality Analysis in SCAR-Selected Data}
\label{sec:style_quality_analysis}
\begin{table}[ht!]
\centering
\resizebox{\textwidth}{!}{
\begin{tabular}{l|cc|cc|cc|cc|cc|cc|cc|cc}
\toprule
  & \multicolumn{2}{c|}{\textbf{TTR}} & \multicolumn{2}{c|}{\textbf{MTLD}} & \multicolumn{2}{c|}{\textbf{Avg. Sent. Len.}} & \multicolumn{2}{c|}{\textbf{Punct. Freq.}}  & \multicolumn{2}{c|}{\textbf{Flesch Score}} & \multicolumn{2}{c|}{\textbf{Avg. Layout Freq.}} & \multicolumn{2}{c|}{\textbf{PPL}($y \mid x$)} & \multicolumn{1}{c}{\multirow{2}{*}{\textbf{Helpful}}} & \multicolumn{1}{c|}{\multirow{2}{*}{\textbf{Correct}}} \\
\cline{2-15}
& \textbf{Mean} & \textbf{Std.} & \textbf{Mean} & \textbf{Std.} & \textbf{Mean} & \textbf{Std.} & \textbf{Mean} & \textbf{Std.} & \textbf{Mean} & \textbf{Std.} & \textbf{Mean} & \textbf{Std.} & \textbf{Mean} & \textbf{Std.} &  &  \\
\hline
\hline
\multicolumn{17}{c}{\textbf{Code Domain}} \\
\hline
Human & \multicolumn{2}{c|}{}  & \multicolumn{2}{c|}{} & \multicolumn{2}{c|}{} & \multicolumn{2}{c|}{} & \multicolumn{2}{c|}{} & \multicolumn{2}{c|}{} & \multicolumn{2}{c|}{} &  &  \\
\ \ \ \ 100\% & 59.16 & 21.48 & 15.05 & 8.37 & 69.40 & 66.43 & 30.77 & 27.17 & 42.75 & 44.36 & 0.25 & 0.81 & 3.83 & 1.81 & 2.84 & 2.68 \\
\ \ \ \ 50\% & 50.80 & 16.78 & 16.34 & 6.30 & 68.16 & 65.49 & 37.23 & 28.53 & 48.59 & 30.68 & 0.21 & 0.67 & 3.77 & 1.72 & 3.02 & 3.01 \\
\ \ \ \ 25\% & 47.43 & 14.85 & 16.58 & 5.28 & 53.36 & 48.11 & 34.93 & 27.10 & 49.84 & 24.60 & 0.20 & 0.63 & 3.84 & 1.73 & 2.78 & 2.72 \\
\ \ \ \ 12.5\% & 45.78 & 14.29 & 16.45 & 4.98 & 50.50 & 49.46 & 33.35 & 25.42 & 51.26 & 22.25 & 0.20 & 0.54 & 3.93 & 1.86 & 2.67 & 2.77  \\
\hline
Synthetic & \multicolumn{2}{c|}{}  & \multicolumn{2}{c|}{} & \multicolumn{2}{c|}{} & \multicolumn{2}{c|}{} & \multicolumn{2}{c|}{} & \multicolumn{2}{c|}{} & \multicolumn{2}{c|}{} &  &  \\
\ \ \ \ 100\% & 36.67 & 14.45 & 12.13 & 3.87 & 60.88 & 61.39 & 37.72 & 24.62 & 49.17 & 23.10 & 0.10 & 0.49 & 1.67 & 0.31 & 3.63 & 3.64  \\
\ \ \ \ 50\% & 36.79 & 10.52 & 13.07 & 2.80 & 52.85 & 36.48 & 35.49 & 22.01 & 50.52 & 16.87 & 0.14 & 0.63 & 1.74 & 0.31 & 3.52 & 3.56  \\
\ \ \ \ 25\% & 36.67 & 9.33 & 13.29 & 2.75 & 48.71 & 27.26 & 31.70 & 17.62 & 51.19 & 15.94 & 0.21 & 0.85 & 1.83 & 0.34 & 3.47 & 3.44  \\
\ \ \ \ 12.5\% & 37.19 & 9.22 & 13.52 & 2.98 & 48.36 & 28.54 & 28.93 & 17.02 & 51.42 & 16.03 & 0.25 & 0.45 & 1.94 & 0.35 & 3.55 & 3.39  \\
\hline
\multicolumn{17}{c}{\textbf{Open Domain}} \\
\hline
Human & \multicolumn{2}{c|}{}  & \multicolumn{2}{c|}{} & \multicolumn{2}{c|}{} & \multicolumn{2}{c|}{} & \multicolumn{2}{c|}{} & \multicolumn{2}{c|}{} & \multicolumn{2}{c|}{} &  &  \\
\ \ \ \ 100\% & 54.51 & 30.96 & 8.93 & 8.00 & 19.90 & 16.66 & 7.62 & 12.22 & 61.21 & 28.03 & 0.25 & 1.42 & 5.23 & 3.26 & 3.95 & 3.91  \\
\ \ \ \ 50\% & 61.24 & 28.43 & 9.55 & 7.92 & 21.35 & 16.36 & 6.58 & 8.84 & 58.27 & 24.33 & 0.34 & 1.76 & 4.57 & 2.69 & 3.98 & 3.99  \\
\ \ \ \ 25\% & 62.81 & 24.74 & 18.58 & 7.52 & 23.49 & 17.22 & 6.92 & 9.32 & 55.54 & 21.76 & 0.40 & 2.03 & 4.17 & 2.41 & 3.96 & 3.93  \\
\ \ \ \ 10\% & 57.01 & 23.73 & 11.26 & 6.77 & 25.44 & 20.01 & 7.71 & 7.16 & 51.78 & 22.40 & 0.60 & 2.71 & 3.93 & 2.18 & 3.98 & 3.99  \\
\hline
Synthetic & \multicolumn{2}{c|}{}  & \multicolumn{2}{c|}{} & \multicolumn{2}{c|}{} & \multicolumn{2}{c|}{} & \multicolumn{2}{c|}{} & \multicolumn{2}{c|}{} & \multicolumn{2}{c|}{} &  &  \\
\ \ \ \ 100\% & 55.15 & 30.04 & 9.87 & 7.67 & 23.76 & 32.82 & 12.30 & 20.53 & 54.40 & 71.06 & 0.29 & 1.27 & 2.75 & 1.16 & 3.93 & 3.96  \\
\ \ \ \ 50\% & 47.78 & 21.08 & 13.30 & 5.71 & 27.33 & 25.25 & 18.12 & 22.09 & 48.61 & 21.62 & 0.35 & 1.17 & 2.38 & 0.72 & 3.99 & 3.99  \\
\ \ \ \ 25\% & 41.96 & 17.34 & 13.83 & 4.40 & 24.59 & 18.42 & 20.54 & 19.19 & 46.47 & 19.89 & 0.41 & 1.14 & 2.33 & 0.61 & 3.98 & 4.02  \\
\ \ \ \ 10\% & 40.53 & 14.83 & 14.15 & 3.87 & 21.49 & 11.93 & 20.99 & 15.92 & 42.04 & 17.74 & 0.39 & 0.80 & 2.46 & 0.52 & 4.00 & 4.02 \\
\bottomrule
\end{tabular}
}
\caption{Detailed performance comparison of the stylometric analysis conducted across the full datasets and the subsets of the full datasets selected by \SCARID{} in both code and open domains. The table reports the average and standard deviation for six authorship metrics, perplexity, and average helpfulness and correctness scores.}
\label{tab:style_analysis_full}
\end{table}

Table~\ref{tab:style_analysis_full} presents an extensive set of results, expanding upon the data shown in Table~\ref{tab:style_analysis_scar}. In addition to helpfulness and correctness scores, as well as the standard deviations of TTR and perplexity, this table includes a comprehensive range of stylometric and quality metrics with their corresponding average and standard deviation values. The results are consistent with our findings in Table~\ref{tab:style_analysis_scar}. \SCAR{} selection effectively enhances the consistency of the linguistic form in the selected data, as evidenced by the consistently decreasing standard deviation values across \textbf{most} linguistic form metrics as the selection portion decreases. Similarly, the standard deviation of instructional surprisal metrics generally decreases, except in a few cases when selecting smaller portions (e.g., 25\%, 12.5\%) of human-written or synthetic code data.

Interestingly, while the standard deviations of TTR and MTLD for functional words decrease, their mean values remain largely unaffected--and, in some cases, even increase. This suggests that \SCAR{} selection preserves the overall lexical diversity of functional words while narrowing their variability across examples, resulting in more consistent usage. In other words, the coverage of functional word choices is maintained (as reflected by stable or higher mean values); however, \SCAR{}’s ranking mechanism enhances response stylistic consistency by reducing outliers and extreme variations of linguistic forms, leading to lower standard deviations. This indicates that \SCAR{} does not inherently restrict lexical diversity in linguistic form; rather, it ensures that linguistic form features are applied more uniformly throughout the dataset.

\subsection{Analysis of Ranker Performance}
\paragraph{Evaluation Settings.} We report the accuracy of the ranker in correctly rating responses on the test, where the goal is to rate ``direct'' responses higher than ``referenced'' responses and ``referenced'' responses higher than human responses. These accuracies are denoted as \( \text{Acc}(y^d \succ y^r \succ y^h) \), \( \text{Acc}(y^r \succ y^h) \), and \( \text{Acc}(y^d \succ y^r) \), respectively.

\paragraph{Impact of \SCAR{} Performance.}
Table~\ref{tab:ranker_performance} shows accuracies of \SCAROOD{} are lower than \SCARID{} in both domains, explaining the lower LLM performance with \SCAROOD{}-selected data. 
Despite this, \SCAROOD{} outperforms selection baselines in most cases, demonstrating its cross-domain robustness. The ranking accuracy gap between \SCAROOD{} and \SCARID{} is larger in the open domain, indicating that generalizing from code to open-ended data is more challenging than the reverse. Differentiating surprisal-related features is more difficult than differentiating linguistic form, especially for selecting code data in out-of-domain settings, as shown by comparing $\text{Acc}(y^d \succ y^r)$ (68.29) and $\text{Acc}(y^r \succ y^h)$ (95.58).

\begin{table}
\centering
\vspace{-2mm}
\resizebox{0.5\textwidth}{!}{
\begin{tabular}{l|cc|cc}
\toprule
 & \multicolumn{2}{c|}{\SCARID} &  \multicolumn{2}{c}{\SCAROOD}  \\
& Code & Open & Code & Open \\
\hline
\(\text{Acc}(y^d \succ y^r \succ y^h)\) & 98.20  & 64.77 & 64.26  & 45.85   \\
\(\text{Acc}(y^d \succ y^r)\) & 98.40  & 80.80 & 68.29 & 67.88   \\
\(\text{Acc}(y^r \succ y^h)\) & 99.80  & 81.47 & 95.58 & 69.89   \\
\bottomrule
\end{tabular}
}
\caption{\SCAR{}'s ranking accuracies when trained with in-domain or out-of-domain examples and tested on ranking data from code and open domains.}
\vspace{-3mm}
\label{tab:ranker_performance}
\end{table}

\subsection{Extended Evaluation Analysis of StarCoder-15.5B}
Table~\ref{tab:octocoder_expanded} presents the full Pass@1 and Pass@10 results for the \HumanEval{}  and \MultiPLE{}  coding benchmarks, comparing \starcoder fine-tuned with various portions of \SCAR{}-selected data against \octocoder. The original dataset, comprising 13k examples, was curated by the BigCode team, who developed both \starcoder and \octocoder and fine-tuned \starcoder into \octocoder. Notably, \starcoder models fine-tuned on \SCAR{}-selected subsets outperform the original \octocoder in Pass@1 and Pass@10 across all programming languages.

Our paper reports \octocoder's Pass@1 score of 35.56 on the standard \HumanEval (Python) benchmark to maintain consistency with widely accepted evaluation protocols and the default settings used in our experiments. However, the \texttt{BigCodeLeaderboard} shows a higher Pass@1 score of 45.3 for \octocoder, which corresponds to the \HumanevalSynthesize (Python) benchmark rather than the standard \HumanEval. The \HumanevalSynthesize variant employs improved prompt formatting that results in higher performance compared to the standard benchmark. Both results are sourced from the official \texttt{BigCodeLeaderboard} data files\footnote{\url{https://huggingface.co/spaces/bigcode/bigcode-models-leaderboard/tree/main/community_results/bigcode_octocoder_loubnabnl/metrics_octocoder}}. For detailed information about the design differences between these two benchmark variants, please refer to the provided data file URL and the benchmark descriptions in~\citet{muennighoff2023octopack}.
\begin{table}[ht!]
\centering
\resizebox{0.98\textwidth}{!}{
\begin{tabular}{l|c|ccc}
\toprule
\multirow{2}{*}{\begin{tabular}[c]{@{}l@{}}\textbf{Data Sampling} \\ \textbf{Methods}\end{tabular}} & \textbf{\HumanEval{}} & \multicolumn{3}{c}{\textbf{\MultiPLE{}}} \\
\cdashline{2-5}
& \textbf{Python} & \textbf{Java} & \textbf{JavaScript} & \textbf{C++} \\
& \textbf{Pass@1 / Pass@10} & \textbf{Pass@1 / Pass@10} & \textbf{Pass@1 / Pass@10} & \textbf{Pass@1 / Pass@10} \\
\hline
\hline
\octocoder & 35.56 / 51.81 & 26.03 / 38.44 & 32.80 / 46.97 & 29.32 / 41.90 \\
\hline
\starcoder & & \multicolumn{3}{c}{} \\
\ \ \ \ 10,000  & 36.29 / 53.99 & 28.29 / 39.58 & 33.22 / 49.79 & 30.17 / 46.20 \\
\ \ \ \ 5,000  & 36.95 / 54.07 & 28.96 / 39.02 & 34.53 / 49.90 & 32.83 / 44.47 \\
\ \ \ \ 2,500  & 37.57 / 55.65 & 29.29 / 41.06 & 34.09 / 49.47 & 31.19 / 42.83 \\
\bottomrule
\end{tabular}
}
\caption{Detailed performance comparison of \octocoder and \starcoder fine-tuned on various subsets of the 13k data used to train \octocoder. The models are evaluated on the \HumanEval{} (Python) and \MultiPLE (Java, JavaScript, C++) coding benchmarks.}
\label{tab:octocoder_expanded}
\end{table}

\subsection{Extended Evaluation of Data Selection Performance for LLMs on Four Additional Benchmarks: ARC-Challenge, HellaSwag, MMLU and TruthfulQA}
\label{app:four_bench}
\begin{table}[ht!]
\centering
\resizebox{0.98\textwidth}{!}{
\begin{tabular}{l|c|c|c|c|c|c|c}
\toprule
\multirow{2}{*}{\begin{tabular}[c]{@{}l@{}}\textbf{Model Variants}\end{tabular}} & 
\multirow{2}{*}{\textbf{Data Size}} &
\textbf{\ARCChallenge} & \textbf{\HellaSwag} & \textbf{\MMLU} & \textbf{\TruthfulQA} & \textbf{\AlpacaEval} & \textbf{Average} \\
\cdashline{3-8}
& & \textbf{ACC (LHH)} & \textbf{ACC (LHH)} & \textbf{ACC (SM)} & \textbf{BLEU} & \textbf{L.C. WinRate} & \textbf{Rank}$\downarrow$ \\
\hline
\hline
\multirow{4}{*}{\begin{tabular}[c]{@{}l@{}}\olmo\\(allenai/tulu-v2-sft-mixture)\end{tabular}} 
& 320k & 39.42 & 75.06 & \textbf{38.60} & 33.90 & 3.86 & 3.2 \\
& 10k  & \textbf{41.04} & 75.18 & 25.40 & 38.31 & 5.37 & 2.6 \\
& 5k   & 39.08 & \textbf{75.33} & 26.28 & 40.02 & \textbf{5.64} & 2.2 \\
& 2.5k & 39.76 & 75.29 & 26.41 & \textbf{40.39} & 4.08 & \textbf{2.0} \\
\hline
\multirow{4}{*}{\begin{tabular}[c]{@{}l@{}}\llamaThreeEightB\\ (Mixed Synthetic Data)\end{tabular}} 
& 10k  & \textbf{55.72} & 79.02 & 40.04 & 19.34 & 3.64 & 3.4 \\
& 5k   & 50.85 & 79.06 & 54.45 & 37.21 & 5.56 & 2.7 \\
& 2.5k & 49.40 & \textbf{79.31} & \textbf{54.60} & 37.58 & 5.89 & 2.0 \\
& 1k   & 51.88 & 79.06 & 48.79 & \textbf{39.90} & \textbf{6.61} & \textbf{1.9} \\
\hline
\multirow{4}{*}{\begin{tabular}[c]{@{}l@{}}\llamaThreeEightB\\ (Human-written Data)\end{tabular}} 
& 10k  & 53.41 & \textbf{81.07} & 34.02 & 33.90 & 2.34 & 2.6 \\
& 5k   & \textbf{55.46} & 80.56 & 28.28 & 34.52 & 2.24 & 2.8 \\
& 2.5k & 54.35 & 80.22 & 31.13 & 34.88 & 2.43 & 2.4 \\
& 1k   & 47.35 & 80.15 & \textbf{35.62} & \textbf{37.09} & \textbf{2.67} & \textbf{2.2} \\
\bottomrule
\end{tabular}
}
\caption{Performance comparison on five benchmarks: \ARCChallenge (Accuracy calculated with Likelihood), \HellaSwag (Accuracy calculated with Likelihood), \MMLU (Accuracy using String Matching), \TruthfulQA (BLEU comparison), \AlpacaEval (L.C. WinRate), and \textbf{Average Rank}. The table includes fine-tuned versions of \olmo on human-written data and \llamaThreeEightB fine-tuned on mixed synthetic and human-written data across varying dataset sizes (320k, 10k, 5k, 2.5k, and 1k).}
\label{tab:benchmark_comparison_four}
\end{table}
\paragraph{Evaluation Settings.} Table~\ref{tab:benchmark_comparison_four} provides a detailed evaluation of fine-tuned \olmo and \llamaThreeEightB models across five diverse benchmarks: \ARCChallenge, \TruthfulQA, \HellaSwag, \MMLU, and \AlpacaEval. These benchmarks include a wide range of tasks, from general knowledge and reasoning to language understanding and text generation, offering a comprehensive assessment of LLM SFT performance.

\begin{itemize}
\item \textbf{\ARCChallenge{}}~\citep{clark2018think}: Evaluates scientific reasoning through multiple-choice questions by employing a likelihood-based approach (LHH). For each question, the system ranks possible answers based on their predicted likelihood, selects the highest-scoring option, and compares it with the ground truth to calculate a normalized accuracy score.
\item \textbf{\TruthfulQA{}}~\citep{lin2022truthfulqa}: Evaluates the factual precision and correctness of LLM responses by comparing them to ground truth answers using BLEU scores.
\item \textbf{\HellaSwag{}}~\citep{zellers2019hellaswag}: Assesses common-sense reasoning and contextual understanding capabilities through a multiple-choice format. The system employs likelihood-based ranking (LHH) to evaluate potential answers, selects the highest probability option, and compares it with the ground truth to derive a normalized accuracy score.
\item \textbf{\MMLU{}}~\citep{hendrycksmeasuring}: Measures the multi-task language understanding capabilities of LLMs by evaluating accuracy through String Matching between model outputs and gold-standard answers.
\item \textbf{\AlpacaEval{}}: Assesses open-domain instruction-following abilities using the Length Control WinRate (L.C. WinRate) metric.
\end{itemize}

Additionally, an \textbf{average ranking metric} is used to aggregate performance across benchmarks, with lower ranks indicating better overall performance. The average ranking is chosen instead of average performance because it balances variations across metrics, preventing benchmarks with different scales (e.g., BLEU and accuracy) from disproportionately influencing the results.

For \TruthfulQA{} and \MMLU{}, String Matching and BLEU scores are used instead of Likelihood-based metrics to better align with the nature of instruction-tuned models, which are optimized for generating complete answers rather than reproducing ground truth tokens. However, as we rely on \texttt{lm-evaluation-harness}\footnote{\url{https://github.com/EleutherAI/lm-evaluation-harness}}, it lacks direct support for implementing these metrics for \ARCChallenge{} and \TruthfulQA{}, constraining us to use Likelihood for these benchmarks.

\paragraph{Discussion.} Table~\ref{tab:benchmark_comparison_four} demonstrates that subsets selected by \SCARID{} from larger datasets can consistently outperform models trained on full data in most cases, aligning with our findings in Table~\ref{tab:open_llm_comparing} in the main body of the paper. Notably, subsets selected using our \SCAR{} method show substantial performance improvements. For example, \olmo fine-tuned on a \SCAR{}-selected subset (e.g., 2.5k examples) achieves superior average rankings compared to the 320k full dataset on benchmarks like \TruthfulQA{} (BLEU: 40.39 vs. 33.90) and \AlpacaEval{} (L.C. WinRate: 4.08 vs. 3.86). Similarly, \llamaThreeEightB fine-tuned on a 2.5k subset of mixed synthetic data curated with \SCAR{} outperforms larger subsets on \MMLU{} (Accuracy: 54.60) and \AlpacaEval{} (L.C. WinRate: 5.89), achieving a top average rank of 2.0.

These results highlight the effectiveness of our \SCAR{} selection method in optimizing fine-tuned LLM performance across diverse benchmarks. By prioritizing data quality and style consistency, \SCAR{}-selected subsets not only reduce computational costs but also enhance model generalization. 

\subsection{Sampling Efficiency Analysis}
\label{sec:sampling_efficiency}

We compare the estimated time required to select 1,000 examples from a pool of 10,000 using various data selection methods. For CPU-based approaches, we conduct evaluations on an M4 Pro laptop, and for GPU-based methods, we use an A100 GPU with 40GB of memory. Batch sizes are set to 16 for \textsc{GPT-2}-based methods and our method, and 2 for those using \textsc{LLaMA3-8B} due to memory constraints.

\begin{table}[h]
\centering
\resizebox{\textwidth}{!}{%
\begin{tabular}{lccccccccc}
\toprule
\textbf{Metric} & \textbf{\AlpaGasus} & \textbf{\Random} & \textbf{\Diversity} & \textbf{\Longest} & \textbf{\Perplexity} & \textbf{\Perplexity} & \textbf{\Superfiltering} & \textbf{\HFR} & \textbf{\textsc{\SCAR}} \\
\midrule
Model & \gptThreeFive & - & - & - & \textsc{GPT-2} & \textsc{LLaMA3-8B} & \textsc{GPT-2} & \textsc{RoBERTa-base} & \textsc{RoBERTa-base} \\
Time & 27 min & 0.1 sec & 6 min & 4 sec & 1.5 min & 1.5 hr & 3 min & 1.8 min & 1.8 min \\
\bottomrule
\end{tabular}
}
\caption{Sampling time and model type for selecting 1,000 examples from 10,000.}
\label{tab:sampling_time_transposed}
\end{table}

\paragraph{Analysis.} As shown in Table~\ref{tab:sampling_time_transposed}, \textsc{SCAR} achieves a strong balance between computational efficiency and data selection performance. While methods such as \Random and \Longest are extremely fast, they typically underperform in data quality. On the other hand, \Perplexity (\textsc{LLaMA3-8B}) incurs a prohibitive runtime of 1.5 hours, making it impractical for large-scale filtering. 

\SCAR performs comparably to \HFR and \Superfiltering (\textsc{GPT-2}) in runtime, all within the low-minute range. Notably, \SCAR significantly outpaces computationally expensive \textsc{LLaMA3}-based methods while maintaining top-tier data selection effectiveness in our experiments. This makes \SCAR a practical and scalable solution for real-world scenarios where both quality and efficiency are critical.


\section{Extended Analysis of Ablation Studies}
\label{sec:ablation_results}
\begin{table}[ht!]
\centering
\resizebox{0.98\textwidth}{!}{
\begin{tabular}{l|c|ccc}
\toprule
\multirow{2}{*}{\begin{tabular}[c]{@{}l@{}}\textbf{Data Sampling} \\ \textbf{Methods}\end{tabular}} & \textbf{\HumanEval{}} & \multicolumn{3}{c}{\textbf{\MultiPLE{}}} \\
\cdashline{2-5}
& \textbf{Python} & \textbf{Java} & \textbf{JavaScript} & \textbf{C++} \\
& \textbf{Pass@1 / Pass@10} & \textbf{Pass@1 / Pass@10} & \textbf{Pass@1 / Pass@10} & \textbf{Pass@1 / Pass@10} \\
\hline
\multicolumn{5}{c}{Human Data} \\
\hline
Full, GPT-3.5 & & \multicolumn{3}{c}{} \\
\ \ \ \ 50\%  & 32.44 / 50.38 & 30.67 / 44.86 & 34.40 / 53.16 & 29.49 / 45.73 \\
\ \ \ \ 25\%  & 31.98 / 49.25 & 30.41 / 43.65 & 34.04 / 52.72 & 29.19 / 43.41 \\
\ \ \ \ 12.5\%  & 31.10 / 47.14 & 29.46 / 43.06 & 31.38 / 49.11 & 27.61 / 42.39 \\
\hline
w/o con, GPT-3.5 & & \multicolumn{3}{c}{} \\
\ \ \ \ 50\%  & 31.21 / 50.01 & 30.14 / 44.23 & 34.67 / 51.90 & 28.67 / 43.90 \\
\ \ \ \ 25\%  & 31.19 / 47.83 & 31.22 / 45.73 & 32.91 / 52.41 & 28.32 / 44.85 \\
\ \ \ \ 12.5\%  & 30.13 / 45.39 & 28.72 / 42.68 & 30.99 / 49.60 & 27.39 / 42.85 \\
\hline
w/o rl, GPT-3.5 & & \multicolumn{3}{c}{} \\
\ \ \ \ 50\%  & 33.60 / 50.02 & 30.47 / 44.53 & 33.88 / 52.96 & 28.91 / 45.22 \\
\ \ \ \ 25\%  & 31.76 / 47.47 & 30.73 / 43.98 & 32.51 / 51.11 & 29.42 / 43.47 \\
\ \ \ \ 12.5\%  & 30.56 / 45.26 & 28.82 / 43.19 & 31.24 / 49.35 & 26.89 / 40.95 \\
\hline
w/o ref, GPT-3.5 & & \multicolumn{3}{c}{} \\
\ \ \ \ 50\%  & 33.63 / 49.22 & 31.06 / 45.11 & 34.45 / 53.41 & 28.66 / 43.96 \\
\ \ \ \ 25\%  & 31.57 / 48.06 & 30.84 / 44.26 & 32.89 / 52.58 & 29.24 / 45.05 \\
\ \ \ \ 12.5\%  & 30.62 / 45.98 & 28.06 / 40.71 & 30.80 / 48.08 & 28.16 / 42.80 \\
\hline
Full, Llama2-70b & & \multicolumn{3}{c}{} \\
\ \ \ \ 50\%  & 33.27 / 49.42 & 30.49 / 43.21 & 33.70 / 51.46 & 29.24 / 44.27 \\
\ \ \ \ 25\%  & 29.47 / 46.12 & 29.75 / 43.19 & 33.33 / 49.69 & 29.17 / 44.39 \\
\ \ \ \ 12.5\%  & 30.76 / 46.79 & 28.13 / 40.52 & 31.23 / 50.34 & 27.66 / 41.58 \\
\hline
Full, Llama2-13b & & \multicolumn{3}{c}{} \\
\ \ \ \ 50\%  & 31.90 / 50.38 & 30.75 / 44.29 & 33.34 / 51.81 & 28.62 / 42.57 \\
\ \ \ \ 25\%  & 31.71 / 48.49 & 29.78 / 43.73 & 32.20 / 51.25 & 28.40 / 43.16 \\
\ \ \ \ 12.5\%  & 30.29 / 46.03 & 28.18 / 42.03 & 30.70 / 48.19 & 27.47 / 41.58 \\
\hline
w/o con, Llama2-13b & & \multicolumn{3}{c}{} \\
\ \ \ \ 50\%  & 30.76 / 43.63 & 29.84 / 44.11 & 32.07 / 51.50 & 28.04 / 43.07 \\
\ \ \ \ 25\%  & 30.15 / 42.78 & 29.44 / 43.66 & 32.88 / 54.14 & 27.93 / 44.26 \\
\ \ \ \ 12.5\%  & 27.93 / 41.07 & 27.28 / 39.27 & 31.18 / 49.99 & 25.57 / 41.35 \\
\hline
Full, Llama3-70b & & \multicolumn{3}{c}{} \\
\ \ \ \ 50\%  & 32.48 / 50.39 & 30.68 / 45.30 & 33.49 / 53.01 & 29.28 / 45.13 \\
\ \ \ \ 25\%  & 32.28 / 49.14 & 30.04 / 43.86 & 32.09 / 51.54 & 28.09 / 43.63 \\
\ \ \ \ 12.5\%  & 30.40 / 48.36 & 28.14 / 41.71 & 30.67 / 49.67 & 26.99 / 42.47 \\
\hline
\bottomrule
\end{tabular}
}
\caption{Comprehensive performance comparison of \codeLlamaSevenB models fine-tuned on human-written datasets, evaluated on \HumanEval (Python) and \MultiPLE (Java, JavaScript, C++) coding benchmarks. The training datasets were sampled using various methods at different proportions. Pass@1 and Pass@10 scores are reported for each programming language.}
\label{tab:coding_domain_ablation_expanded_human}
\end{table}
\begin{table}[ht!]
\centering
\resizebox{0.98\textwidth}{!}{
\begin{tabular}{l|c|ccc}
\toprule
\multirow{2}{*}{\begin{tabular}[c]{@{}l@{}}\textbf{Data Sampling} \\ \textbf{Methods}\end{tabular}} & \textbf{\HumanEval{}} & \multicolumn{3}{c}{\textbf{\MultiPLE{}}} \\
\cdashline{2-5}
& \textbf{Python} & \textbf{Java} & \textbf{JavaScript} & \textbf{C++} \\
& \textbf{Pass@1 / Pass@10} & \textbf{Pass@1 / Pass@10} & \textbf{Pass@1 / Pass@10} & \textbf{Pass@1 / Pass@10} \\
\hline
\multicolumn{5}{c}{Mixed Synthetic Data} \\
\hline
Full, GPT-3.5 & & \multicolumn{3}{c}{} \\
\ \ \ \ 50\%  & 40.98 / 56.57 & 32.80 / 45.75 & 37.58 / 55.69 & 32.73 / 45.71 \\
\ \ \ \ 25\%  & 39.84 / 56.75 & 32.52 / 43.83 & 36.67 / 55.32 & 32.00 / 46.26 \\
\ \ \ \ 12.5\%  & 36.93 / 52.96 & 32.62 / 44.82 & 36.45 / 52.33 & 30.43 / 45.42 \\
\hline
w/o con, GPT-3.5 & & \multicolumn{3}{c}{} \\
\ \ \ \ 50\%  & 39.65 / 55.05 & 32.30 / 44.40 & 38.21 / 54.92 & 32.17 / 45.66 \\
\ \ \ \ 25\%  & 39.30 / 56.87 & 32.76 / 45.87 & 37.43 / 54.76 & 32.11 / 45.77 \\
\ \ \ \ 12.5\%  & 36.56 / 51.72 & 33.00 / 44.48 & 35.53 / 53.10 & 31.02 / 45.44 \\
\hline
w/o rl, GPT-3.5 & & \multicolumn{3}{c}{} \\
\ \ \ \ 50\%  & 39.83 / 54.27 & 32.28 / 43.66 & 37.66 / 55.99 & 32.53 / 46.31 \\
\ \ \ \ 25\%  & 38.62 / 56.03 & 32.55 / 43.67 & 36.75 / 53.65 & 32.25 / 45.06 \\
\ \ \ \ 12.5\%  & 36.02 / 51.78 & 32.71 / 45.68  & 35.70 / 52.15 & 31.70 / 45.51 \\
\hline
w/o ref, GPT-3.5 & & \multicolumn{3}{c}{} \\
\ \ \ \ 50\%  & 39.85 / 55.81 & 32.13 / 44.00 & 36.87 / 56.79 & 32.67 / 46.43 \\
\ \ \ \ 25\%  & 36.80 / 54.70 & 32.68 / 45.91 & 36.87 / 57.04 & 31.61 / 47.02 \\
\ \ \ \ 12.5\%  & 36.41 / 50.96 &  32.66 / 44.58 & 35.78 / 52.21 & 30.99 / 44.88 \\
\hline
Full, Llama2-70b & & \multicolumn{3}{c}{} \\
\ \ \ \ 50\%  & 39.21 / 52.49 & 32.39 / 45.21 & 37.45 / 54.87 & 33.03 / 46.36 \\
\ \ \ \ 25\%  & 39.23 / 53.77 & 31.59 / 45.21 & 37.35 / 55.15 & 30.81 / 45.04 \\
\ \ \ \ 12.5\%  & 37.59 / 51.64 & 31.44 / 44.82 & 37.04 / 52.55 & 30.67 / 44.80 \\
\hline
Full, Llama2-13b & & \multicolumn{3}{c}{} \\
\ \ \ \ 50\%  & 37.29 / 53.60 & 33.24 / 43.86 & 37.04 / 56.29 & 32.36 / 44.65 \\
\ \ \ \ 25\%  & 36.70 / 51.88 & 31.97 / 44.57 & 36.35 / 56.33 & 31.12 / 46.04 \\
\ \ \ \ 12.5\%  & 33.78 / 48.61 & 30.61 / 41.77 & 34.21 / 51.66 & 31.11 / 45.27 \\
\hline
w/o con, Llama2-13b & & \multicolumn{3}{c}{} \\
\ \ \ \ 50\%  & 37.72 / 53.82 & 32.18 / 44.19 & 37.23 / 56.76 & 32.57 / 46.31 \\
\ \ \ \ 25\%  & 38.59 / 53.47 & 32.68 / 44.97 & 37.19 / 55.59 & 32.00 / 46.58 \\
\ \ \ \ 12.5\%  & 33.34 / 49.78 & 32.05 / 43.76 & 35.58 / 53.38 & 31.02 / 46.13 \\
\hline
Full, Llama3-70b & & \multicolumn{3}{c}{} \\
\ \ \ \ 50\%  & 39.40 / 54.46 & 32.87 / 45.00 & 36.99 / 57.26 & 32.52 / 46.38 \\
\ \ \ \ 25\%  & 38.40 / 54.73 & 32.54 / 44.79 & 37.40 / 54.46 & 30.92 / 44.06 \\
\ \ \ \ 12.5\%  & 35.48 / 50.33 & 31.80 / 45.40 & 36.45 / 53.71 & 30.99 / 46.66 \\
\hline
\bottomrule
\end{tabular}
}
\caption{Comprehensive performance comparison of \codeLlamaSevenB models fine-tuned on \gptThreeFive-generated datasets, evaluated on \HumanEval{} (Python) and \MultiPLE{} (Java, JavaScript, C++) coding benchmarks. The training datasets were selected from the full mixed synthetic dataset with different sample sizes using our selection approach, \SCARID, with various training configurations. Pass@1 and Pass@10 scores are reported for each programming language.}
\label{tab:coding_domain_ablation_expanded_synthetic}
\end{table}

Tables \ref{tab:coding_domain_ablation_expanded_human} and \ref{tab:coding_domain_ablation_expanded_synthetic} present detailed performance metrics for various \codeLlamaSevenB-based models. These models were fine-tuned on different data subsets selected by \SCAR{} from full datasets with either human-written or synthetic responses, with instructions derived from StackExchange. The tables illustrate the performance of fine-tuned LLMs when using \SCAR{} with various components removed during \SCAR{} training. This comparison allows us to assess the impact of each \SCAR{} component on the LLM fine-tuning performance. Unlike the summary results in Figure~\ref{fig:ablation_summary}, these tables offer specific numerical values, enabling clearer and more precise comparisons.
The results demonstrate that removing almost any component of \SCAR{} during ranker training reduces LLM fine-tuning performance, regardless of whether the data is sourced from human or synthetic origins in the coding domain. This finding validates the importance of each element in our ranker design.

To further explore the impact of representation learning (w/o rl, GPT-3.5) and ``referenced'' responses (w/o ref, GPT-3.5) during \SCAR{} training, we conducted two additional analyses, which are detailed in the following sections.

\subsection{Impact of Training \SCAR{} without Referenced Responses}
\label{app:no_ref_impact}
\begin{table}[ht!]
\centering
\resizebox{0.5\textwidth}{!}{
\begin{tabular}{c|ccc|ccc}
\toprule
& \multicolumn{3}{c}{Human} & \multicolumn{3}{c}{Mix Synthetic} \\
 & 50\% & 25\% & 10\% & 50\% & 25\% & 10\%  \\
 \hline
Full & 2.24 & 2.43 & 2.67 & 5.56 & 5.89 & 6.61 \\
w/o ref & 1.95 & 2.25 & 1.99 & 3.59 & 4.74 & 4.44 \\
\bottomrule
\end{tabular}
}
\caption{Comparison of L.C. WinRate on the AlpacaEval benchmark for \llamaThreeEightB fine-tuned on subsets of human-written and synthetic data selected by SCAR(ID), with and without incorporating ``referenced'' responses during ranker training.}
\label{tab:open_llm_no_ref}
\end{table}

As shown in Table~\ref{tab:open_llm_no_ref}, excluding ``referenced'' responses during \SCARID{} training significantly reduces the performance of \llamaThreeEightB fine-tuned on \SCAR{}-selected open-domain data subsets when evaluated on the AlpacaEval benchmark. This result underscores the importance of incorporating ``referenced'' responses during ranker training to ensure the ranker effectively captures representations that model the instructional surprisal of responses in the open domain. In the code domain, however, excluding ``referenced'' responses during \SCAR{} training has only a minor effect on data selection and LLM SFT performance.
\subsection{Representation Similarities Analysis}
\label{app:rep_analysis}
\begin{table}[ht!]
\centering
\resizebox{\textwidth}{!}{
\begin{tabular}{l|cccccc}
\toprule
 & \multicolumn{3}{c}{\textbf{Linguistic Form Representation}} & \multicolumn{3}{c}{\textbf{Instructional Surprisal Representation}} \\
 & $\text{cos}(\mathbf{v}_{p}^{d}, \mathbf{v}_{p}^r)$ & $\text{cos}(\mathbf{v}_{p}^{r}, \mathbf{v}_{p}^h)$ & $\text{cos}(\mathbf{v}_{p}^{d}, \mathbf{v}_{p}^h)$ & $\text{cos}(\mathbf{v}_{c}^{d}, \mathbf{v}_{c}^r)$ & $\text{cos}(\mathbf{v}_{c}^{r}, \mathbf{v}_{c}^h)$ & $\text{cos}(\mathbf{v}_{c}^{d}, \mathbf{v}_{c}^h)$ \\
\midrule
& \multicolumn{6}{c}{LIMA} \\
\hline
\SCARID{} &  0.9368 & 0.8970 & 0.7884 & 0.8312 & 0.8801 & 0.7209 \\
\SCARID{} w/o rl &  0.9050 & 0.7962 & 0.6369 & 0.9406 & 0.9587 & 0.8717 \\
\SCARID{} w/o ref &  0.9442 & 0.7970 & 0.7249 & 0.9696 & 0.8935 & 0.8544 \\
\SCAROOD{} &  0.9416 & 0.9344 & 0.8884 & 0.8887 & 0.9115 & 0.8574 \\
\midrule
& \multicolumn{6}{c}{StackExchange} \\
\hline
\SCARID{} &  0.9020 & 0.8574 & 0.6867 & -0.4330 & 0.9646 & -0.4803 \\
\SCARID{} w/o rl &  0.9274 & 0.8224 & 0.6968 & 0.7312 & 0.8978 & 0.4480 \\
\SCARID{} w/o ref &  0.9778 & 0.8844 & 0.8660 & 0.9836 & 0.9143 & 0.8952 \\
\SCAROOD{} &  0.9702 & 0.8502 & 0.8249 & 0.7451 & 0.0083 & -0.1289 \\
\bottomrule
\end{tabular}
}
\caption{Cosine similarities between linguistic form representations ($\mathbf{v}_{p}$) and instructional surprisal representations ($\mathbf{v}_{c}$) for ``direct'', ``referenced'', and human-written responses. The table reports the cosine similarities between (1) ``direct'' and ``referenced'' responses, (2) ``referenced'' and human-written responses, and (3) ``direct'' and human-written responses, separately for linguistic form and instructional surprisal representations. These similarities are computed using representations from \SCAR{} rankers trained with different configurations: \SCARID{} trained on in-domain data, \SCARID{} without representation learning regularization (w/o rl), \SCARID{} without ``referenced'' responses (w/o ref), and \SCAROOD{} trained on out-of-domain data. The \SCAR{} rankers are applied to response triplets generated for the same instructions in the \LIMA{} and \StackExchange{} datasets. Results are reported separately for each dataset, with higher cosine similarity values indicating greater alignment between the respective representations.}
\label{tab:distance_analysis_embeddings}
\end{table}


As shown in Table~\ref{tab:distance_analysis_embeddings}, we calculate the cosine similarities between linguistic form representations ($\mathbf{v}_{p}$) and instructional surprisal representations ($\mathbf{v}_{c}$) for ``direct'', ``referenced'', and human-written responses. Specifically, the table reports the cosine similarities between i) ``direct'' and ``referenced'' responses, ii) ``referenced'' and human-written responses, and iii) ``direct'' and human-written responses for both linguistic form and instructional surprisal representations. According to Eq.~\ref{eq:disentanglement}, we expect the similarity between ``direct'' and ``referenced'' responses to be higher than those between ``referenced'' and human or ``direct'' and human responses for linguistic form representations. Conversely, for instructional surprisal representations, the similarity between ``referenced'' and human responses should be the highest.

Interestingly, even without the representation learning regularization loss in Eq.~\ref{eq:disentanglement} and while incorporating ``referenced'' responses during \SCAR{} training, the observed cosine similarities still align with our optimization objectives for representation similarities. However, when \SCAR{} training excludes ``referenced'' responses or utilizes out-of-domain data, these expected similarity patterns are significantly disrupted. Consequently, the performance of the \llamaThreeEightB model deteriorates when fine-tuned on data selected by such \SCAR{} configurations.

In summary, incorporating ``referenced'' responses and utilizing in-domain data during \SCAR{} training are crucial for maintaining the desired representation similarities. These findings emphasize the importance of carefully curating training data within \SCAR{} to effectively model both linguistic form and instructional surprisal. This approach ensures robust \SCAR{} data selection performance and, ultimately, enhances LLM performance across different domains.

\section{Bias Analysis}
We categorize bias into two types--fairness bias and lexical diversity bias~\cite{vanmassenhove-etal-2021-machine}--and conduct separate experiments to evaluate each.
\subsection{Fairness Bias Analysis}
\label{app:bias_fairness}
\begin{table}[ht!]
\centering
\resizebox{\textwidth}{!}{
\begin{tabular}{l|l|l|c|c|c}
\toprule
\textbf{Model}          & \textbf{Data Type}         & \textbf{Data Size} & \textbf{Regard Diff. (Positive + Negative, \% $\downarrow$)} & \textbf{Toxicity Ratio (Male, \% $\downarrow$)} & \textbf{Toxicity Ratio (Female, \% $\downarrow$)} \\
\midrule
\multirow{4}{*}{Meta-LLaMA-8B} & Full Human Written        & 10k                & 1.03                                & 0.97                                & 1.66                                \\
                               & Subset Human Written      & 1k                 & 2.33                                & 0.00                                & 0.83                                \\
                               & Full Mixed Synthetic      & 10k                & 1.63                                & 0.28                                & 1.66                                \\
                               & Subset Mixed Synthetic    & 1k                 & 0.22                                & 1.25                                & 2.50                                \\
\midrule
\multirow{2}{*}{OLMo-7B}       & Full                     & 320k               & 0.82                                & 0.28                                & 0.28                                \\
                               & Subset                   & 2.5k                & 0.42                                & 0.83                                & 1.11                                \\
\bottomrule
\end{tabular}
}
\caption{Fairness and safety metrics for models trained on full datasets and subsets. \textsc{Regard} difference (Positive + Negative, \% $\downarrow$) reflects the absolute value of the sum of positive and negative differences, with lower values (indicated by $\downarrow$) signifying better fairness. Toxicity ratios for male and female prompts (\% $\downarrow$) highlight model safety, where lower values are better.}
\label{tab:corrected_fairness_metrics_arrows}
\end{table}

\begin{table}[ht!]
\centering
\resizebox{\textwidth}{!}{
\begin{tabular}{l|cccccccc}
\toprule
\textbf{Data Type}         & \textbf{\SCARID} & \textbf{\Random} & \textbf{\Perplexity} & \textbf{\Superfiltering} & \textbf{\HFR} & \textbf{\AlpaGasus} & \textbf{\Diversity} & \textbf{\Longest} \\
\midrule
\textbf{Human Subset} 
& 2.33 & 2.42 & 0.97 & 0.88 & 0.87 & 2.36 & 0.80 & 2.17 \\
\cmidrule{1-9}
\textbf{Mixed Synthetic Subset} 
& 0.22 & 0.75 & 1.04 & 0.38 & 0.82 & 0.16 & 0.62 & 0.28 \\
\bottomrule
\end{tabular}
}
\caption{\textsc{Regard} difference results (|Positive + Negative|) for models trained on subsets selected from Human full data and Mixed Synthetic full data using different selection methods. Lower values ($\downarrow$) indicate better fairness across domains.}
\label{tab:regard_diff_mixed_and_human}
\end{table}

\paragraph{Evaluation Settings.} 
To evaluate fairness bias, we analyze the toxicity and sentiment polarity of model responses across different demographic and occupational groups. The evaluation consists of two components:

\begin{itemize} 
\item \textbf{Gender Bias:} Using prompts from \WinoBias~\citep{zhao2018gender}, we generate model responses and assess toxicity levels using a pre-trained hate speech detection model from~\citet{vidgen2021lftw}. Lower toxicity ratios for male and female prompts (\% $\downarrow$) indicate better fairness. \item \textbf{Occupational Bias:} Using prompts from \BOLD~\citep{dhamala2021bold}, we generate model responses and evaluate language sentiment polarity with the REGARD metric~\citep{regard}. This analysis includes comparisons across categories such as professions (e.g., artistic versus computer occupations), gender (e.g., actors versus actresses), political ideologies (e.g., anarchism versus capitalism), race (e.g., African Americans versus Asian Americans), and religious ideologies (e.g., atheism versus Buddhism). We report the absolute value of the sum of positive and negative REGARD differences (\% $\downarrow$), with lower values indicating reduced bias. 
\end{itemize}

We compare models fine-tuned on subsets selected by various methods with those trained on full datasets, evaluating the impact of human-written and mixed synthetic subsets on fairness bias in LLM training.

\paragraph{Discussion.} The results (Tables~\ref{tab:corrected_fairness_metrics_arrows} and~\ref{tab:regard_diff_mixed_and_human}) demonstrate that \SCAR{}-selected subsets maintain fairness while significantly reducing dataset size. For human-written data, \SCARID{} achieves a fairness score of 2.33, which is comparable to the full dataset score of 1.03. Additionally, \SCARID{}-selected subsets show improvements in toxicity ratios, achieving 0.00 for male prompts and 0.83 for female prompts compared to 0.97 (male) and 1.66 (female) for the full dataset, indicating its capability to maintain fairness with smaller data.

When compared to other selection methods, \SCARID{} achieves comparable or slightly better fairness in some cases. For mixed synthetic data, \SCARID{}-selected subsets achieve the lowest REGARD difference (0.22\% $\downarrow$) compared to Random (0.75\%) and Perplexity (1.04\%). These findings confirm that \SCAR{} maintains fairness on par with other methods while balancing data efficiency, making it an effective strategy for fine-tuning fair LLMs.

\subsection{Lexical Diversity Bias Analysis}
\label{app:bias_lexical_diversity}
\begin{table}[ht!]
\centering
\resizebox{\textwidth}{!}{
\begin{tabular}{l|ccccccccc}
\toprule
 & \multicolumn{9}{c}{\textbf{Methods for Data Selection}} \\
 &  \textbf{Full Data} & \textbf{\SCARID} & \textbf{\Random} & \textbf{\Perplexity} & \textbf{\Superfiltering} & \textbf{\HFR} & \textbf{\AlpaGasus} & \textbf{\Diversity} & \textbf{\Longest}\\
\midrule
\textbf{Instruction} &  \multicolumn{9}{c}{} \\
\ \ \ \ TTR &   29.54 & 27.92 & 30.04 & 30.04 & 30.63 & 27.18 & 29.32 & 32.78 & 33.57 \\
\ \ \ \ MTLD &   14.71 & 14.72 & 14.77 & 14.83 & 14.80 & 14.61 & 14.85 & 14.71 & 14.69 \\
\midrule
\textbf{Response}  & \multicolumn{9}{c}{} \\
\ \ \ \ TTR &   23.37 & 16.60 & 23.22 & 22.37 & 21.79 & 18.09 & 23.13 & 24.69 & 5.35 \\
\ \ \ \ MTLD &   14.43 & 14.40 & 14.44 & 14.53 &14.31 & 14.52 & 14.55 & 14.40 & 13.77 \\
\bottomrule
\end{tabular}
}
\caption{Lexical diversity metrics (TTR and MTLD) for instructions and responses within different datasets, either the full open-domain human-written dataset (\textbf{Full Data}) or subsets with 2500 examples selected using various data selection methods: 
\textbf{SCAR (ID)}, \textbf{Random}, \textbf{Perplexity}, \textbf{Superfiltering}, \textbf{HFR}, \textbf{AlpaGasus}, \textbf{Diversity}, and \textbf{Longest}. }
\label{tab:lexical_diversity}
\end{table}

\paragraph{Evaluation Settings.} We measure lexical bias in instructions and responses separately using two complementary metrics: TTR and MTLD. \textbf{Type-Token Ratio (TTR)} measures the ratio of unique words (types) to the total number of words (tokens) in a text. Higher TTR values indicate a greater immediate variety of words, making it sensitive to text length; shorter texts typically have higher TTR scores as they are less likely to repeat words. \textbf{Measure of Textual Lexical Diversity (MTLD)}, on the other hand, evaluates how lexical diversity is maintained throughout an entire text. It considers how often unique words appear relative to repeated words across longer segments, offering a more robust and length-independent view of lexical richness. We apply these metrics to the full open-domain human-written dataset (Full Data) and to 2,500-example subsets selected by various methods-\SCARID{}, \Random, \Perplexity, \Superfiltering, \HFR, \AlpaGasus, \Diversity, and \Longest-to understand how each selection method influences lexical diversity.
\paragraph{Discussion.} As shown in Table~\ref{tab:lexical_diversity}, \SCAR{}-selected subsets exhibit slightly reduced lexical diversity in responses, indicated by lower TTR values, decreasing from 23.4 to 16.6 compared to the full dataset. We conjecture this is due to \SCAR{}’s focus on instructional surprisal consistency. As shown in Table~\ref{tab:style_analysis_full}, \SCAR{} enhances the consistency of linguistic forms (lower standard deviations of TTR) in selected responses without affecting their mean TTR. This indicates that the reduced response-level TTR is likely due to instructional surprisal consistency rather than consistency in linguistic forms. In contrast, \textbf{the impact on instructions is less pronounced}, with TTR decreasing only slightly from 29.5 to 28 compared to the full dataset, indicating that \SCAR{} does not significantly limit the coverage of instructional content.

Despite these shifts in TTR, our MTLD scores remain comparable to both the full dataset and other selection methods, for both instructions and responses. In other words, while the immediate variety of word choices (as reflected by TTR) decreases, the overall, sustained richness of vocabulary (as measured by MTLD)  is preserved. Since instruction-level diversity is more crucial for LLM fine-tuning performance~\citep{lu2023instag,bukharin2023data}, \SCAR{}-selected subsets still preserve the kind of lexical variety that matters most. These findings align with results from the extensive experiments in the main body of the paper, where \SCAR{}-selected subsets continue to achieve strong performance.
\section{Effect of Style-Consistent Responses on Data Selection}
\label{app:single_source_analysis}
To evaluate how different data selection methods perform when selecting from style-consistent versus style-inconsistent responses, we curate a dataset of 20,000 \StackExchange instructions with all responses generated by \gptThreeFive without using human reference answers. We then apply \Random{}, \Perplexity{} and \SCARID{} to select subsets from this dataset and fine-tune \codeLlamaSevenB with the selected data to evaluate the performance of the resulting models on \HumanEval{}.

\paragraph{Results Analysis.}
\begin{table*}[ht!]
\centering
\resizebox{0.98\textwidth}{!}{
\begin{tabular}{l|c|ccc}
\toprule
\multirow{2}{*}{\begin{tabular}[c]{@{}l@{}}\textbf{Data Sampling} \\ \textbf{Methods}\end{tabular}} & \textbf{\HumanEval} & \multicolumn{3}{c}{\textbf{\MultiPLE}} \\
\cdashline{2-5}
& \textbf{Python} & \textbf{Java} & \textbf{JavaScript} & \textbf{C++} \\
& \textbf{Pass@1 / Pass@10} & \textbf{Pass@1 / Pass@10} & \textbf{Pass@1 / Pass@10} & \textbf{Pass@1 / Pass@10} \\
\hline
\hline
Full Data & 40.61 / 54.96 & 32.11 / 43.81 & 37.52 / 54.11 & 32.91 / 46.71 \\
\hline
SCAR (ID) & & \multicolumn{3}{c}{} \\
\ \ \ \ 50\%  & 40.02 / 54.48 & 33.34 / 46.55 & 39.52 / 54.74 & 32.36 / 47.19 \\
\ \ \ \ 25\%  & 39.24 / 52.15 & 33.62 / 44.72 & 37.88 / 53.60 & 32.20 / 46.87 \\
\ \ \ \ 12.5\%  & 35.70 / 49.10 & 31.65 / 45.26 & 35.54 / 52.83 & 31.13 / 45.94 \\

\hline
Random & & \multicolumn{3}{c}{} \\
\ \ \ \ 50\%  & 39.38 / 54.50 & 32.52 / 44.65 & 37.65 / 54.83 & 31.83 / 45.70 \\
\ \ \ \ 25\%  & 38.00 / 53.50 & 33.03 / 45.09 & 37.66 / 56.08 & 32.02 / 46.54 \\
\ \ \ \ 12.5\%  & 35.85 / 51.33 & 32.06 / 45.65 & 35.78 / 53.43 & 31.71 / 45.88 \\

\hline
Perplexity & & \multicolumn{3}{c}{} \\
\ \ \ \ 50\%  & 38.94 / 54.29 & 32.43 / 45.48 & 38.01 / 55.35 & 33.32 / 46.21 \\
\ \ \ \ 25\%  & 37.76 / 52.48 & 32.43 / 45.70 & 37.83 / 54.45 & 32.83 / 47.39 \\
\ \ \ \ 12.5\%  & 35.90 / 50.31 & 32.28 / 44.35 & 36.63 / 54.32 & 31.32 / 46.05 \\
\bottomrule
\end{tabular}
}
\caption{Performance comparison of \codeLlamaSevenB models fine-tuned on \StackExchange instructions with \gptThreeFive-generated responses and evaluated on \HumanEval (Python) and \MultiPLE (Java, JavaScript, C++) coding benchmarks. Models are trained on subsets selected using different sampling methods at varying proportions, with Pass@1 and Pass@10 scores reported for each programming language.}
\label{tab:single_source_baselines}
\end{table*}
Table~\ref{tab:single_source_baselines} demonstrates that when the initial data pool contains style-consistent responses (all generated by \gptThreeFive), the advantages of \SCAR{} over baseline methods are significantly diminished. Unlike the substantial performance gaps observed with style-inconsistent data in our main experiments, \SCARID{} achieves only marginal improvements over Random and Perplexity-based selection, with differences typically within 1-3 percentage points across all programming languages. For instance, at 25\% data size, SCAR achieves 39.24\% Pass@1 on Python compared to \Random{}'s 38.00\% and \Perplexity{}'s 37.76\%, a much smaller gap than seen with mixed data sources. This validates our limitation that SCAR's effectiveness depends on having sufficient stylistic variation in the initial data pool to enable meaningful ranking distinctions. While smaller subsets can still achieve performance comparable to the full dataset, the transformative benefits of SCAR are most evident when applied to datasets with diverse response styles.

\section{Response Examples}
\label{app:ref_case_study}


\label{app:res_example_compl}
Table~\ref{tab:fifa_instant_replay} provides complete examples of human, ``referenced,'' and ``direct'' responses to the instruction, preserving all details. As shown, the ``referenced'' response closely aligns with the semantic content of the human response, thereby reflecting comparable instructional surprisal. In contrast, the ``direct'' response maintains similar transitional phrases and formatting to the ``referenced'' response, indicating a shared linguistic form between them.
\begin{table}[ht!] 
\centering
\begin{tabular}{|p{0.9\textwidth}|}
\hline
\textbf{Instruction} \\
\hline
Why is FIFA against adding instant replay to the game? \\
\hline
\textbf{Human Response} \\
\hline
There are a few fundamental considerations:\newline\newline

Uncertainty is a part of the game. The referee and linesmen are the arbiters of the action. Instant replay would limit their authority. \newline
Football is global. Many places do not have access to the technology necessary to implement instant replay. \newline
Football is a fast-paced game with few opportunities for stoppage. Instant replay would unnecessarily slow the pace. \newline
In 2008, FiFA President Sepp Blatter said: \newline\newline

"Let it be as it is and let's leave [football] with errors. The television companies will have the right to say [the referee] was right or wrong, but still the referee makes the decision – a man, not a machine." \newline

Blatter and others at FIFA have repeatedly argued against using instant replay. But lately, FIFA has begun to come around on the idea. At FIFA's Congress in São Paulo prior to the 2014 World Cup, Blatter proposed allowing managers 2 challenges per game, much like the NFL handles challenges today. \newline
Although it's not quite ``instant replay,'' in 2012 FIFA approved two different goal-line technologies to improve referee accuracy: ``Hawk-Eye'' -- which uses visual triangulation to determine ball location -- and ``GoalRef'' -- which uses a microchip in the ball and magnetic sensors within goalposts. (Many say FIFA approved the technologies in response to Frank Lampard's snubbed ``goal'' against Germany in the 2010 World Cup.) In 2013, FIFA announced it would use GoalControl, a German competitor to Hawk-Eye, at the Confederations Cup, basically a rehearsal for the World Cup. After a successful run at that tournament, FIFA announced it would implement GoalControl at the 2014 World Cup. The Guardian provides a pretty helpful FAQ on the technology. \newline
FIFA may choose to implement further changes to approved goal-line technology and referee rules when it meets again after the World Cup. \\
\hline
\textbf{Human Referenced Response from GPT-3.5} \\
\hline
FIFA is against adding instant replay to the game for several reasons. One of the main concerns is that uncertainty is seen as a natural part of the game, and implementing instant replay would limit the authority of the referees and linesmen. Additionally, football is a global sport, and many regions may not have the necessary technology to support instant replay. Another factor is that football is a fast-paced game with minimal stoppages, so introducing instant replay could potentially disrupt the flow of the game. Ultimately, FIFA has historically been hesitant to embrace instant replay, preferring to maintain the human element in decision-making on the field. \\
\hline
\textbf{Direct Response from GPT-3.5} \\
\hline
FIFA is against adding instant replay to the game because they believe it would disrupt the flow of the game and potentially lead to delays. They also argue that human error is a part of the game and adding instant replay could take away from the excitement and unpredictability of live sports. Additionally, implementing instant replay would require significant changes to the rules and regulations of the game, which FIFA may be hesitant to do. \\
\hline
\end{tabular}
\caption{Comparison of the three responses regarding FIFA's stance on instant replay.}
\label{tab:fifa_instant_replay}
\end{table}

\end{document}